\documentclass[journal,twoside]{IEEEtran}

\usepackage{cite}

\ifCLASSINFOpdf
  \usepackage[pdftex]{graphicx}
\else
  \usepackage[dvips]{graphicx}
\fi
\usepackage{placeins}

\usepackage{amsmath}
\usepackage{amssymb}  
\usepackage{mathtools}

\usepackage{algorithm} 
\usepackage{algorithmic}
\usepackage{array}
\usepackage[caption=false,font=footnotesize]{subfig}
\usepackage{url}
\usepackage{siunitx}
\usepackage{hyperref}
\usepackage{multirow}
\usepackage{threeparttable}
\usepackage[multiple]{footmisc}

\usepackage{color}
\usepackage[dvipsnames]{xcolor}
\usepackage{verbatim}

\newcommand{\smallsurr}[3]{\underset{\begin{subarray}{c} s \sim d^{#1} \\ a \sim #2 \end{subarray}}{\mathbb{E}} \left[ #3 \right]}
\newcommand{\surrA}[3]{\frac{1}{1-\gamma}\underset{\begin{subarray}{c} s \sim d^{#1} \\ a \sim #2 \end{subarray}}{\mathbb{E}} \left[ #3 \right]}

\newcommand{\E}{\textrm{E}}

\begin{document}
\title{Not Only Rewards But Also Constraints: Applications on Legged Robot Locomotion}

\author{Yunho~Kim,
        Hyunsik~Oh,
        Jeonghyun~Lee,
        Jinhyeok~Choi,
        Gwanghyeon~Ji, \\
        Moonkyu~Jung,
        Donghoon~Youm,
        and~Jemin~Hwangbo$^{*}$
\thanks{This work was supported by Samsung Research Funding \& Incubation Center for Future Technology at Samsung Electronics under Project Number SRFC-IT2002-02.}
\thanks{All authors are with Robotics and Artificial Intelligence Lab, KAIST, Daejeon, South Korea}
\thanks{$*$ corresponding author \tt\footnotesize jhwangbo@kaist.ac.kr}
}


\maketitle

\begin{abstract}
Several earlier studies have shown impressive control performance in complex robotic systems by designing the controller using a neural network and training it with model-free reinforcement learning. However, these outstanding controllers with natural motion style and high task performance are developed through extensive reward engineering, which is a highly laborious and time-consuming process of designing numerous reward terms and determining suitable reward coefficients. In this work, we propose a novel reinforcement learning framework for training neural network controllers for complex robotic systems consisting of both \textit{rewards} and \textit{constraints}. To let the engineers appropriately reflect their intent to constraints and handle them with minimal computation overhead, two constraint types and an efficient policy optimization algorithm are suggested. The learning framework is applied to train locomotion controllers for several legged robots with different morphology and physical attributes to traverse challenging terrains. Extensive simulation and real-world experiments demonstrate that performant controllers can be trained with significantly less reward engineering, by tuning only a single reward coefficient. Furthermore, a more straightforward and intuitive engineering process can be utilized, thanks to the interpretability and generalizability of constraints. The summary video is available at \url{https://youtu.be/KAlm3yskhvM}.

\end{abstract}




\section{Introduction}
\FloatBarrier

\IEEEPARstart{R}{ecently}, learning-based methods have gained significant popularity for designing controllers in complex robotic systems. These techniques employ a neural network as a controller, mapping the robot's observations to control inputs \cite{hwangbo2019actuator}. The network parameters are trained using either expert demonstration data (imitation learning) \cite{pan2017drivingimitation, zhao2023manipulationimitation, peng2018deepmimic} or interaction data (reinforcement learning) \cite{hwangbo2017drone, lee2020blind, andrychowicz2020dex}. Since expert demonstration data is often limited for robotic systems compared to readily available interaction data from both physics simulations and real-world experiments, reinforcement learning has become a dominant approach. These methods offer distinct advantages over model-based approaches, especially in scenarios where the system's complexity increases with more available contacts (e.g., legged robots, high-degree-of-freedom robot hands) and environmental uncertainties grow due to factors like sensor noise and disturbances.

\begin{figure}[t!]
\centering
\includegraphics[width=\columnwidth]{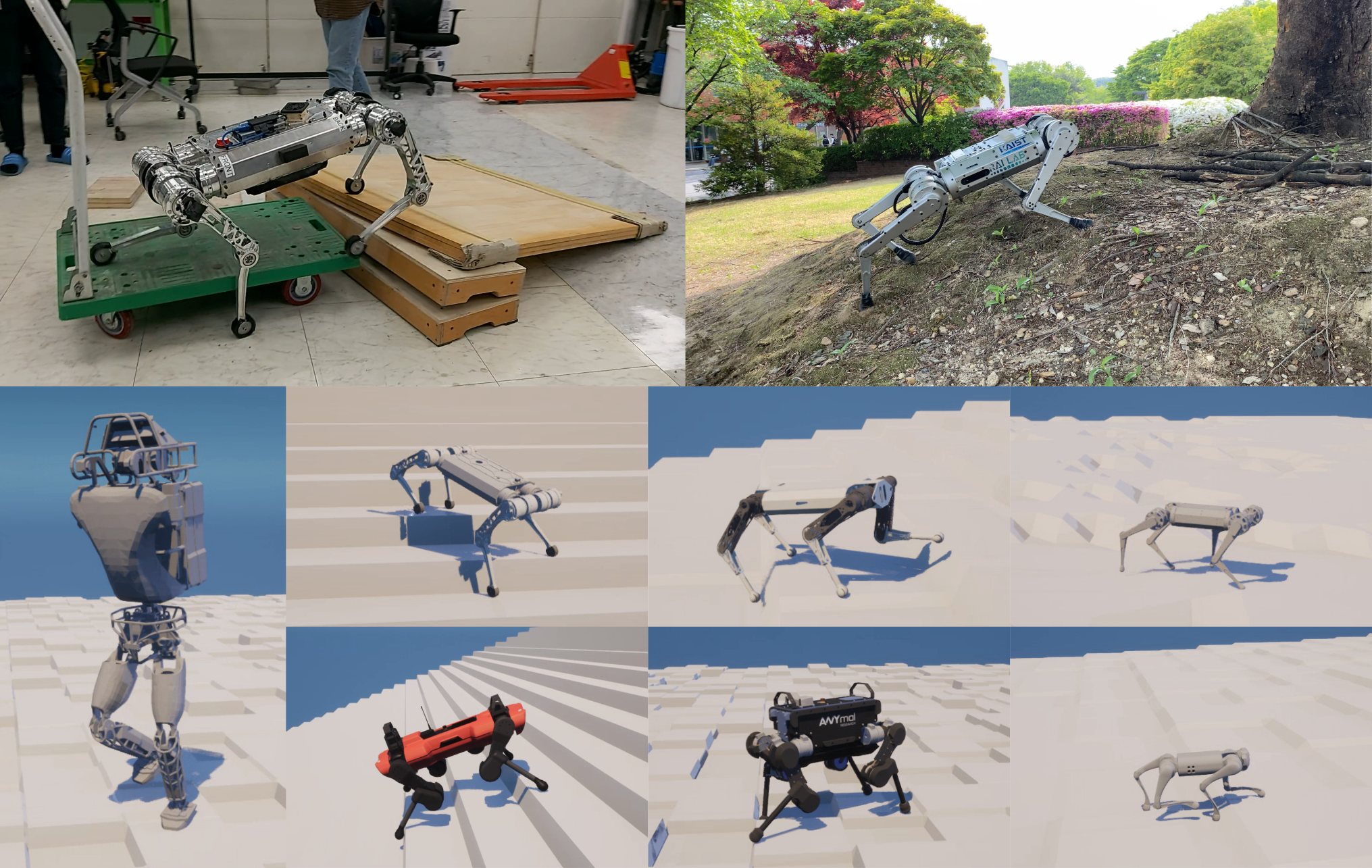}
\caption{Several legged robots trained with our learning framework to locomote in challenging terrains. The robots include six quadrupedal robots and one bipedal robot.}
\label{fig:multiple_robot}
\vspace*{-0.25cm}
\end{figure}

When engineers build a neural network controller with reinforcement learning, they typically do the following steps: first, design a neural network architecture with inductive bias based on the observation and action spaces; second, generate abundant environment interaction scenarios (e.g., random terrains for legged robot locomotion, diverse object meshes for robot hand grasping); and third, design reward terms and tune their reward coefficients until a high-performing controller is achieved, that satisfies the robot's physical constraints (e.g., joint torque and velocity limits) and exhibits a natural motion style. Among the three steps, the last one is the most time-consuming because the tuning process has to be repeated several times. Finding the relative weights for each of the reward terms based on the resulting robot's motion is not trivial because there are often more than ten reward terms \cite{ji2022concurrent, choi2023deformable}.

In this work, we want to raise the following fundamental question: \textit{Why constraints have not been used explicitly to train policies for complex robotic systems?} Reinforcement learning can be thought of as a numerical optimization problem: finding the neural network parameters that maximize the objective, which is the weighted sum of the reward terms. As the system becomes more complex and the desired motions become more agile, it is inevitable to put more effort into reward engineering to make the objective surface smooth with fewer local optima and guide the optimization toward the desired point. However, in numerical optimization problems, constraints are also utilized to narrow down the solution search space to the region that is desirable for the engineer or feasible in the actual system. 

Similarly, if constraints can be explicitly defined in the reinforcement learning framework, instead of relying solely on rewards, there can be several advantages. First, the training pipeline will be more generalizable across similar robot platforms. If the learning framework is only designed with rewards, the performance of the controller can vary across multiple robots due to the differences in feedback signals for each reward component. These variations result in different control behaviors, making the controller's performance unique to each robot's specific characteristics and dynamics. On the other hand, constraints can be used as generalizable conditions that the controller should satisfy even when the robot changes. For instance, when we include a trotting gait pattern constraint in the training of locomotion controllers, it can be applied consistently to all quadruped robots, irrespective of their size, mass, and morphology. Second, the engineering process will be more straightforward and less time-consuming. There will be fewer reward terms to adjust, as the rewards that previously acted as soft penalties to accommodate the robot's hardware limits or define its motion style are no longer needed. Instead, they are defined as constraints to drive the policy towards the approximate desired region that coincides with the engineer's intent. Furthermore, constraint parameters and limits can be set more intuitively, compared to reward coefficients that correspond to the relative importance of each term, because they have physical meanings (e.g., joint angle limits) and can even be set automatically from the robot description files (e.g., URDF files).

To this end, we propose a reinforcement learning framework for complex articulated systems consisting of both rewards and constraints. Appropriate constraint types are suggested, where each can be utilized in a way that suits the engineer's intent. Inspired by previous works in the constrained reinforcement learning literature (also known as safe reinforcement learning), an efficient policy optimization algorithm is suggested to search for a policy that maximizes the reward while satisfying multiple constraints with a negligible amount of additional computation cost. The proposed learning framework is used to train controllers for legged robot locomotion in challenging terrains, which was possible only with considerable reward engineering. Through extensive experiments conducted in both simulation and the real world, involving multiple robots with different morphologies (e.g., leg numbers, mechanical leg designs) and properties (e.g., robot mass, actuator parameters) as shown in Figure \ref{fig:multiple_robot}, we demonstrate the effectiveness of our framework. Specifically, we show that high-performance controllers can be obtained with significantly less reward engineering (i.e., only three reward coefficients and a single reward coefficient modification) and by exploiting more generalizable and straightforward constraints. 

Unlike other related works \cite{gangapurwala2020gcpo, lee2023evaluationCRL} that focused on safety for the benefit of constraints, our work investigates a new advantage of utilizing constraints from a practical engineering perspective. To the best of our knowledge, our work is the first to show highly robust simulation and real-world robot performance for multiple legged robots (i.e., six quadrupedal robots and one bipedal robot) using constrained reinforcement learning and demonstrate the benefits of constraints from an engineering standpoint.

\subsection{Contributions}
The primary contribution of this paper is to provide new perspectives on designing a learning-based controller for complex articulated systems. Our experiments demonstrate how using constraints makes the engineering process more efficient, compared to relying solely on well-tuned rewards to reflect the engineer's intent. This improvement stems from the generalizability and interpretability of constraints, along with the reduced need for extensive reward designs. The concrete contributions are elaborated on below:
\begin{itemize}
    \item We introduce a reinforcement learning framework for complex articulated systems consisting of both \textit{rewards} and \textit{constraints}, drawing inspiration from existing works in constrained reinforcement learning. Our approach incorporates suitable constraint types and an efficient policy optimization algorithm. The framework is designed to be highly scalable in terms of the number of constraints while enabling high-performance policy learning.
    \item Through extensive simulation and real-world experiments involving multiple legged robots with varying morphologies and physical properties (i.e., six quadrupedal robots and one bipedal robot), we demonstrate the capability of leveraging constraints in the learning pipeline to improve robot generalizability and engineering intuitiveness, which are difficult to achieve when relying solely on rewards, while maintaining highly robust control performance.
    \item Building on our encouraging experimental results, we present an intriguing direction for considering constraints as a new tool that can be used in conjunction with or instead of rewards to design neural network controllers. This alleviates the burden of tedious reward engineering, which is a major bottleneck for robot control with reinforcement learning.
\end{itemize}

\subsection{Outline}
The rest of the paper is structured as follows. After Sec.~\ref{sect:related_work} reviewing relevant literature, Sec.~\ref{sect:background} provides short background information about constrained reinforcement learning. Sec.~\ref{sect:framework} explains the proposed learning framework including constraint types and the policy optimization algorithm. Sec.~\ref{sect:legged_robot} demonstrates how we applied the proposed framework for legged robot locomotion. Sec.~\ref{sect:experiment_results} illustrates the implementation details and experimental results. Finally, Sec.~\ref{sect:conclusion_future_work} concludes the paper with a discussion and future research directions.

\section{Related Work}
\label{sect:related_work}

\subsection{Control of complex robotic systems}
In model-based control literature, engineers typically do the following to design a controller for a system: first, analyze the system's kinematic and dynamic equations; second, define heuristics or objectives and constraints that represent the desired motion; and third, design a control system with numerical optimization or simpler techniques (e.g., Proportional-Integral-Differential control) and generate control inputs. These approaches are effective when the system is relatively simple and the environment is almost deterministic. However, they become challenging when the system becomes more complex and the environmental uncertainties grow. 

To handle these challenges, learning-based control methods with deep neural networks are actively applied to controlling complex articulated systems. The learning-based control literature can be subdivided into two categories based on the data used for training the neural network: imitation learning and reinforcement learning. Imitation learning trains the network to mimic the expert's behavior through expert demonstration data. The network can be trained directly via behavior cloning with data aggregations \cite{pan2017drivingimitation} or indirectly by matching the expert's state distribution \cite{peng2018deepmimic, peng2021amp}. However, demonstration data for complex articulated systems is often limited or not available. Reinforcement learning is an appealing alternative in these cases. It aims to train a controller that maximizes the expected reward sum by using interaction data, rather than expert demonstration data, which can be readily obtained in both simulation and the real world. Neural networks can be used to parameterize the dynamic model of the environment (model-based reinforcement learning) \cite{nagabandi2020dex, kim2022nav} or the controller itself that maps observations to control inputs (model-free reinforcement learning). In this work, we are mostly focused on model-free reinforcement learning (RL) methods, that have been successfully applied to controlling various complex systems including animated characters \cite{heess2017deepmindhumanoid}, high-dof robot hands \cite{andrychowicz2020dex}, drones \cite{hwangbo2017drone}, and mobile manipulators \cite{fu2022legmanip} with well-engineered dense reward signals.

\subsection{Legged robot locomotion}
Legged robots have a great chance of navigating a variety of rough terrains that wheeled robots find challenging. They may be deployed anywhere that people and animals can go and traverse the region by carefully selecting footholds and altering the base motions. However, the complexity of the hardware and the underactuated nature makes it difficult to design a control algorithm for these hardware platforms. Trajectory optimization techniques are a widely used methodology in the domain of model-based control. In these methods, the control is separated into two modules: planning and tracking. The planning module generates base motion and foot trajectories using numerical optimization or heuristics to satisfy the given high-level commands from the user, such as a velocity command and a desired foot contact sequence. The tracking module then generates actuation torques to follow the plan. Although the effectiveness of these methods has been demonstrated using physical robots for both blind locomotion \cite{bellicoso2018dynamic, di2018cheetahmpc} and perceptive locomotion \cite{jenelten2022tamols, grandia2023perceptive}, there are challenges associated with their application in wild environment settings due to the environmental uncertainties, sensor noise, and unforeseen corner cases.

Model-free deep reinforcement learning (RL) has recently risen to attention as an alternative method for legged locomotion. These techniques use neural networks to model a locomotion controller, and its parameters are automatically trained to maximize the designed reward signals. Because RL requires lots of data to get superior control performance, a recent trend is to train the controller in the physics simulator \cite{hwangbo2019actuator, lee2020blind, miki2022perception, rudin2022issac} and deploy it in the real world (i.e., sim-to-real). Techniques including domain randomization \cite{peng2018domainrandom}, domain adaptation, and hybrid simulation \cite{hwangbo2019actuator} are suggested to bridge the gap between the simulation and the real world. Previous research on quadruped robots demonstrated that RL controllers may be very reliable and generalize to difficult contexts by simulating a large amount of related experiences. In particular, Lee et al. \cite{lee2020blind} and Kumar et al. \cite{kumar2021rma} demonstrated robust blind locomotion capabilities in a variety of hard conditions, including slippery and rocky hills, through teacher-student learning in multiple simulated rigid terrains.  Choi et al. \cite{choi2023deformable} demonstrated high-speed quadrupedal locomotion in a variety of deformable terrains, such as soft beach sand, using efficient granular media simulation. Miki et al. \cite{miki2022perception} and Agarwal et al. \cite{agarwal2023egovision} introduced improved RL locomotion controllers by exploiting exteroceptive information such as terrain geometries around each foot or front-facing depth camera data. Furthermore, RL controllers demonstrated high-speed dynamic locomotion of small-scale quadruped robots \cite{ji2022concurrent, margolisyang2022rapid}, multiple gait transitions \cite{shao2021multiple, kim2021multiple}, human-size bipedal robot locomotion \cite{radosavovic2023humanoid}, and even more agile movements like parkour \cite{hoeller2023anymalparkour} or environment interactions \cite{cheng2023legsasmanip}.

Although the advantages of using model-free deep reinforcement learning for legged robot locomotion have been shown in various previous works in perspective of the control performance, less attention is given to how the controllers are engineered. To achieve such a robust controller with natural and smooth joint movements, a variety of reward terms (often ten or more) are designed first, and the reward coefficients for each of the terms are extensively tuned until satisfactory performance is achieved. This reward engineering process is the most time-consuming part of designing the controller (This procedure often takes several months). Finding the relative weights for each of the reward terms based on the resulting robot's motion is not trivial and it highly depends on the engineer's experience and intuition.

There are some previous works to reduce the burden of reward engineering. Escontrela et al. \cite{escontrela2022adversarialAnymal}, Wu et al. \cite{wu2023adverserialTO}, and Fuchioka et al. \cite{fuchioka2023opt} utilized an auxiliary task to imitate the natural motions present in expert demonstration data, derived either from animal motions or trajectory optimization. However, the motions that the controller can generate are limited to the available expert demonstration data. Feng et al. \cite{feng2023genloco} proposed a morphology randomization technique to generate a general controller for quadruped robots and remove the robot-specific engineering process. However, training a single controller for a broad range of quadruped robot morphologies results in a conservative controller and is a well-known problem of domain randomization techniques. Furthermore, their experiments were restricted to indoor settings with simple behaviors (e.g., go forward, turn in place), limiting their ability to demonstrate high controller robustness.

In this work, we propose a learning framework that is formulated with both rewards and constraints. Because constraints are generalizable, interpretable, and can be utilized to guide the policy to the desired region, our method reduces the burden of extensive reward engineering by requiring fewer reward terms (only three) to tune and making the training process more general across similar robot platforms.

\subsection{Constrained reinforcement learning}
In reinforcement learning, the agent learns to maximize the expected reward sum through trial and error. In many realistic settings, however, giving an agent complete freedom may be unacceptable. For example, consider a legged robot learning to move forward with minimal torque usage. It may be undesirable for the robot to encounter body contact rather than the feet (e.g., self-collision, contact between the ground and the robot trunk) or to utilize a joint velocity range close to the hardware limit. In such cases, guiding the search space to the desirable region is required, and a natural way to do this is via constraints.

The Constrained Markov Decision Process (CMDP) framework \cite{altman2021cmdp} is a well-known and well-researched model for reinforcement learning with constraints, where agents must learn to maximize reward while satisfying multiple constraints defined as the expectation of the cost terms. We recommend readers who are not familiar with the CMDP framework or constrained reinforcement learning to first read Sec. \ref{sect:background}, which explains them in detail.

Constrained reinforcement learning methods can be subdivided into two categories: \textit{optimization criterion} and \textit{exploration process} \cite{garcia2015saferl_survey}. The optimization criterion methods propose policy update rules to satisfy constraints, and the exploration process methods utilize additional safe policies to intervene during the exploration when the agent's state is unsafe. There are several prior works on robotics leveraging exploration process methods for safe exploration when directly trained in real-world \cite{yang2022safeexplore, bharadhwaj2020safeexplore}. In this work, we are focused on optimization criterion methods because safe exploration is not important when the robot is trained in the simulation. The optimization criterion methods can be further divided into two categories based on how the policy is updated: \textit{lagrangian} and \textit{trust-region} methods.

Lagrangian methods treat the constrained problem as an unconstrained problem by inducing Lagrange multipliers and solving it using primal-dual methods \cite{chow2017pdo, tessler2018rcpo}. Although constraints are mostly satisfied when the policy converges, they are sensitive to hyperparameters (e.g., initialization of Lagrange multipliers) and often show unstable training \cite{achiam2017cpo}. Trust-region methods update the policy through linear approximation of the objective within the trust region. This subfield contains a variety of algorithms, which can be further subdivided into first-order methods \cite{liu2020ipo, zhang2022p3o, xu2021crpo} and higher-order methods \cite{achiam2017cpo, yang2020pcpo}. Only those algorithms that are directly relevant to our work will be covered in greater detail. Achiam et al. \cite{achiam2017cpo} proposed Constrained Policy Optimization (CPO) by extending TRPO to the CMDP framework with linear approximations of constraints. Although monotonic policy improvements with constraint satisfaction are empirically shown, the requirement of the second-order derivatives imposes implementation difficulty and computation burden. First-order methods alleviate this problem by converting the constrained problem into an unconstrained problem and using first-order gradients to optimize the policy. Different kernel functions are used for stable constraint satisfaction. Liu et al. \cite{liu2020ipo} proposed Interior-point Policy Optimization (IPO), which uses logarithmic barrier functions for the constraint functions to optimize the policy within the constraint-satisfying region. Yang et al. \cite{zhang2022p3o} proposed Penalized Proximal Policy Optimization (P3O), which incorporates clamping functions for the constraint functions to impose a linear cost gradient penalty up until the constraint is satisfied and ignore the cost gradient inside the constraint-satisfying area.

Our policy optimization algorithm is based on IPO but several improvements are added to make the algorithm scalable for multiple constraints. First, an adaptive constraint thresholding method is used to appropriately set the constraint limits based on the current policy's performance and guide the policy to the constraint-satisfying region using steep gradients of logarithmic barrier functions. Second, a multi-head cost value function is proposed to parallelize cost value function training and cost advantage computation in a memory-efficient manner. With these improvements, we demonstrate that our framework is capable of training controllers for a variety of legged robots, which necessitates the incorporation of more than ten constraints. Furthermore, we showcase the robustness of these controllers in real-world scenarios. Our strong experimental results demonstrate that our method is superior for real-world robotic applications when compared to the prior works \cite{achiam2017cpo}, \cite{liu2020ipo} that were tested on simulated benchmark tasks \cite{ray2019safetyGym} with one or two constraints. There is a previous work by Gangapurwala et al. \cite{gangapurwala2020gcpo} that defined the problem of training locomotion controllers for legged robots in a CMDP formulation. However, their framework required cost coefficient engineering because they handled several constraints as rewards with appropriately set coefficients. Furthermore, their learning framework required considerable reward engineering due to multiple reward terms and a reference trajectory generator, whereas our framework requires only three reward terms and no demonstration data. Our comparison experiment with \cite{gangapurwala2020gcpo} also demonstrates that the controller obtained with our method is superior in the perspective of safety and task performance.

\section{Background}
\label{sect:background}

\subsection{Constrained Markov Decision Process}
Constrained Markov Decision Process (CMDPs) \cite{altman2021cmdp} is an augmented MDP framework that is used to define a constrained reinforcement learning problem. A CMDP is defined as $(S$, $A$, $P$, $R$, $C_{1,..,K}$, $\rho$, $\gamma)$, where $S$ denotes the state space, $A$ denotes the action space, $P:S\times A\times S \mapsto \mathbb{R}^{+}$ is the transition model, $R:S\times A \mapsto \mathbb{R}$ is the reward function, $C_k:S\times A \mapsto \mathbb{R}$ is the cost function for $\forall k\in\{1,..., K\}$, $\rho$ is the initial state distribution, and $\gamma$ is the discount factor. Let $J(\pi)$ and $J_{C_k}(\pi)$  denote the expected discounted return of policy $\pi$ with respect to the reward and cost functions as defined as follows:
\begin{equation}
\label{eq:reward_cost_objectives}
\begin{gathered}
J(\pi) := \underset{\rho, \pi, P}{\mathbb{E}}\left[\sum_{t=0}^\infty\gamma^tR(s_t, a_t, s_{t+1})\right], \\
J_{C_k}(\pi) := \underset{\rho, \pi, P}{\mathbb{E}}\left[\sum_{t=0}^\infty\gamma^tC_k(s_t, a_t, s_{t+1})\right].
\end{gathered}
\end{equation}
The constrained reinforcement learning problem can be defined as finding a policy that maximizes $J(\pi)$ and satisfies all constraints $J_{C_k}(\pi)$ as follows:
\begin{equation}
\label{eq:constrained_rl_problem}
\begin{gathered}
\pi^{*} = \arg \max_{\pi \in \Pi_{\theta}} \; J(\pi) \\
\text{s.t. } J_{C_k} (\pi) \leq d_k \;\;\; \forall k \in \{1, ..., K\},\\
\end{gathered}
\end{equation}
where $d_k$ is the threshold for the $k$th constraint, $\Pi_{\theta}$ is the set of parameterized policies with parameters $\theta$. The set of policies that satisfy all constraints is denoted as the \emph{feasible region}. According to the derivation done by Schulman et al. \cite{schulman2015trpo} and Achiam et al. \cite{achiam2017cpo}, the constrained reinforcement learning problem (\ref{eq:constrained_rl_problem}) can be approximated inside the trust region as below:
\begin{equation}\label{eq:constrained_rl_problem_approx}
\begin{gathered}
\pi_{i+1} = \arg \max_{\pi \in \Pi_{\theta}} \;   \smallsurr{\pi_i}{\pi}{A^{\pi_i}(s,a)}  \\
\text{s.t. }  J_{C_k} (\pi_i) + \surrA{\pi_i}{\pi}{A^{\pi_i}_{C_k} (s,a)} \leq d_k \;\;\; \forall k\\
 \bar{D}_{KL} (\pi || \pi_i) \leq \delta,
\end{gathered}
\end{equation}
where $d^{\pi}$ is the discounted state distribution of policy $\pi$, defined by $d^{\pi} (s) = (1-\gamma) \sum_{t=0}^{\infty} \gamma^t P(s_t = s | \pi)$, $A^{\pi}$ is the reward advantage function, $A^{\pi}_{C_k}$ is the \textit{k}th cost advantage function, $\bar{D}_{KL}(\pi || \pi_i) = \E_{s\sim d^{\pi_i}}\left[D_{KL}(\pi||\pi_i)[s] \right]$, and $\delta > 0$ is the maximum step size. The optimal safe policy for the approximated constrained reinforcement learning problem (\ref{eq:constrained_rl_problem_approx}) can be obtained by various algorithms \cite{achiam2017cpo, yang2020pcpo, liu2020ipo, zhang2022p3o, xu2021crpo}.

\section{Learning Framework}
\label{sect:framework}
\begin{figure*}[t]
\centering
\includegraphics[width=\linewidth]{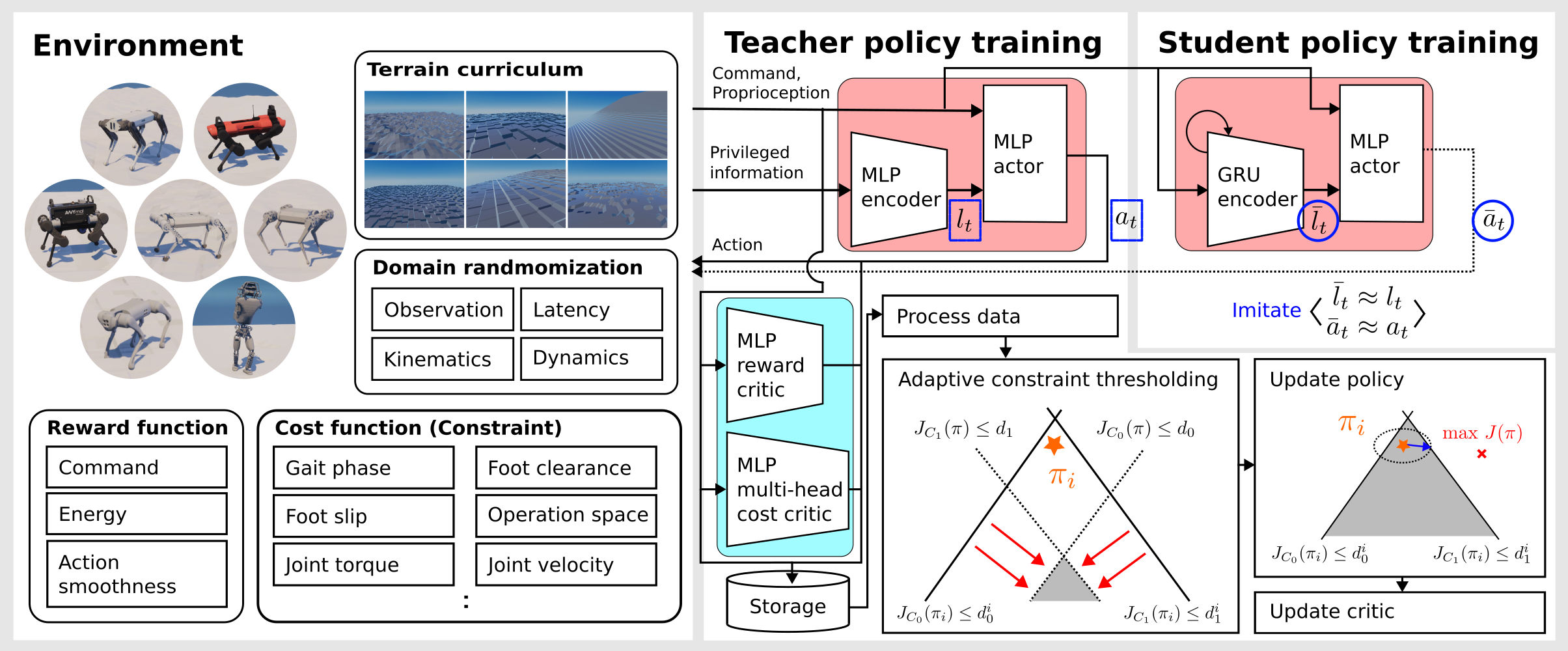}
\caption{Overall framework that leverages both rewards and constraints to train locomotion controllers for various legged robots}
\label{fig:learning_framework}
\vspace*{-0.25cm}
\end{figure*}

\subsection{Constraint Types}
The policy search space is typically too large when training a neural network policy from scratch to control a complex articulated system, making it difficult to design a policy that yields desired motions. Designing advanced reward signals with fine-grained reward engineering can be a solution \cite{hwangbo2019actuator, lee2020blind, miki2022perception, choi2023deformable}, but the process is highly time-consuming and robot-specific. In this work, we rather leverage constraints to restrict the search space and train the policy within the constraint-satisfying regions.
 
In reinforcement learning, stochastic policies are often used \cite{schulman2015trpo, schulman2017ppo} for better exploration. Thus, constraints for the policies should be defined in a probabilistic manner, which is the biggest difference from deterministic constraints commonly used in numerical optimization problems. We use two types of constraints: \textit{probabilistic} and \textit{average} constraints. Each constraint can be customized in a form that suits the engineer's intent.

\subsubsection{Probabilistic Constraints}
A probabilistic constraint is used to limit the probability of an undesirable event. It can be set by defining the cost function as an indicator function as below:
\begin{equation}
\label{eq:probabilistic_constraint_cost_func}
    C_k(s, a, s') = 
\begin{cases}
    0, & \text{if } (s, a, s') \in \mathbf{S}\\
    1,              & \text{otherwise,}
\end{cases}
\end{equation}
where \textbf{S} is the robot's desirable event space.
Then the constraint is written as follows:
\begin{equation}
\label{eq:probabilistic_constraint}
\mathrm{Prob}((s, a, s') \notin \mathbf{S})
=\underset{\rho, \pi, P}{\mathbb{E}}\left[C_k(s, a, s')\right] \leq D_k,
\end{equation}
where $D_k \in [0, 1]$ is the probability threshold for the $k$th constraint. If $D_k$ is set to zero, the policy is trained to completely stay away from encountering the undesirable event. Consequently, the probabilistic constraint can be employed to establish constraints for the undesirable region that the robot should avoid at every time step. For example, a robot's center of mass position should remain within a specific distance threshold from the ground to pass through the overhanging obstacles. We can define the desired area $\textbf{S}$ for the center of mass and set the limit of the probability constraint to zero. However, satisfying the constraint of $D_k = 0$ in stochastic MDP settings is challenging. Thus, in our work, $D_k$ is instead set to a fairly small value.

\subsubsection{Average Constraints}
An average constraint is used to restrict the average of some physical variables of the robot to be below a desired threshold. It can be set by defining the cost function as below:
\begin{equation}
\label{eq:average_constraint_cost_func}
    C_k(s, a, s') = f(s, a, s'), \\
\end{equation}
where $f(s, a, s')$ is the corresponding physical variable.
Then the constraint is written as
\begin{equation}
\label{eq:average_constraint}
\underset{\rho, \pi, P}{\mathbb{E}}\left[f(s, a, s')\right] 
=\underset{\rho, \pi, P}{\mathbb{E}}\left[C_k(s, a, s')\right] \leq D_k,
\end{equation}
where $D_k$ is the value threshold for the $k$th constraint. This type is appropriate when the goal is to limit the average value rather than to enforce constraints at every step. For instance, it is preferable for legged robots to have low foot velocities when making contact with the ground to avoid slippage. Because the foot velocity may not be zero on slippery surfaces, it is undesirable to use a probabilistic constraint. Instead, we can specify a threshold below which the average foot contact velocity should be.

\subsection{Policy optimization}
The probabilistic and average constraints are handled as discounted cumulative constraints during the policy optimization as follows:
\begin{equation}
J_{C_k}(\pi) := \underset{\rho, \pi, P}{\mathbb{E}}\left[\sum_{t=0}^\infty\gamma^tC_k(s_t, a_t, s_{t+1})\right] \leq d_k,
\end{equation}
where $d_k = D_k / (1 - \gamma)$ is the modified constraint limit. This has been done in previous works \cite{achiam2017cpo, liu2020ipo} to efficiently optimize the policy in CMDP formulation (\ref{eq:constrained_rl_problem}) with a near-constraint satisfaction guarantee \cite{achiam2017cpo}.

We use the Interior-point Policy Optimization (IPO) \cite{liu2020ipo} to solve the constrained reinforcement learning problem (\ref{eq:constrained_rl_problem}). IPO converts the constrained problem into an unconstrained problem using logarithmic barrier functions as follows:
\begin{equation}
\label{eq:ipo_problem}
\underset{\pi \in \Pi_\theta}{\mathrm{maximize}} \; J(\pi) + \sum_{k=1}^K\log{\left(d_k - J_{C_k}(\pi)\right)}/t,
\end{equation}
where $t > 0$ is the hyperparameter that determines the steepness of the logarithmic barrier functions. With some approximations (\ref{eq:constrained_rl_problem_approx}), the IPO objective (\ref{eq:ipo_problem}) is modified as follows:
\begin{equation}
\label{eq:ipo_problem_approx}
\begin{gathered}
\underset{\pi \in \Pi_\theta}{\mathrm{maximize}} \; 
 \smallsurr{\pi_i}{\pi}{A^{\pi_i}(s,a)} \\ 
+ \sum_{k=1}^K\log(d_k - (J_{C_k} (\pi_i) + \surrA{\pi_i}{\pi}{A^{\pi_i}_{C_k} (s,a)})) / t \\
\text{s.t. }  \bar{D}_{KL} (\pi || \pi_i) \leq \delta.
\end{gathered}
\end{equation}
In the original paper \cite{liu2020ipo}, the IPO objective (\ref{eq:ipo_problem_approx}) is optimized with Proximal Policy Optimization (PPO) \cite{schulman2017ppo}. However, we utilized Trust Region Policy Optimization (TRPO) \cite{schulman2015trpo} because the optimizer's explicit line search allows for more stable policy improvement and constraint feasibility checking \footnote{We also experimented with the PPO\cite{schulman2017ppo} optimizer and found several algorithmic advantages, such as ease of implementation and the availability to utilize recurrent neural networks, albeit with the trade-off of training stability.}. The advantage functions $A^{\pi}$ and $A^{\pi}_{C_k}$ are computed using Generalized Advantage Estimation (GAE) \cite{schulman2015gae} and neural network value functions. Each advantage function then undergoes additional standardization and zero-mean normalization before computing the policy gradients.

Because we are utilizing constrained reinforcement learning to train controllers for complex robotic systems, the method should be capable of dealing with a large number of constraints. However, earlier publications in the literature, such as CPO and IPO, had limits when it came to explicitly addressing these issues. To that purpose, we incorporate two modifications to the IPO to make it scalable for the number of constraints.

\subsubsection{Adaptive constraint thresholding}
IPO requires the policy to be located in the feasible region during the training because it uses logarithmic barrier functions. However, this setting is not practical in situations where policies are randomly initialized. For these circumstances, a naive approach \cite{xu2021crpo} can be utilized, in which one of the violated constraints is chosen and the policy is adjusted until it enters the feasible region by directly reducing the cost advantage functions. This, however, makes the training inefficient because updating the violated constraints one at a time becomes difficult as the number of constraints increases.

In this work, we widen the feasible region by adaptively adjusting the constraint limit based on the performance of the current policy $\pi_i$. Specifically, the $k$th constraint limit for the $i$th policy update $d^i_k$ is set as 
\begin{equation}
\label{eq:adaptive_constraint_threshold}
d^i_k = max(d_k, J_{C_k} (\pi_i) + \alpha \cdot d_k), \\
\end{equation}
where $J_{C_k} (\pi_i) = {\underset{\rho, \pi_i, P}{\mathbb{E}} \left[ C_k(s, a, s') \right]} / (1 - \gamma)$ and $\alpha$ is the hyperparameter that determines how much the feasible region will be enlarged. 

This makes the problem always feasible due to zero-mean normalization done for the cost advantage function $A^{\pi}_{C_k}$ and brings a similar effect as scale normalization by parameterizing the enlarging range with $d_k$. Furthermore, the steep inclination of a barrier function gradually guides the policy toward the desired constraint region defined with threshold $d_k$.

\subsubsection{Multi-head cost value function}
For each constraint, parameterized cost value functions (in our case, neural networks) are required to compute the cost advantage function with GAE \cite{schulman2015gae}. When there are multiple constraints, one implementation choice is to use independent neural networks with no shared parameters for each cost value function. However, the number of parameters to train increases linearly with the number of constraints. Instead, we employ a multi-head cost value function, in which a single neural network predicts all cost values rather than just one. In this setting, all cost-value functions share the same neural network backbone, and only the output layer dimension grows when more constraints are introduced.

The overall policy optimization procedure to solve the reinforcement learning problem with multiple constraints (\ref{eq:constrained_rl_problem}) is summarized in Algorithm \ref{alg:modified_IPO}.
\begin{algorithm}[t]
\caption{Modified Interior-point Policy Optimization} 
\label{alg:modified_IPO}
\hspace*{0.01in} {\bf Input:} 
Policy $\pi^{\theta_0}$, Value function $V^{\phi_0}$, and Multi-head Cost Value function $V^{\psi_0}_{C_{1,..K}}$. Hyperparameter $t$ for logarithmic barrier functions and $\alpha$ for adaptive constraint thresholding.\\
\hspace*{0.01in} {\bf Output:}
Policy $\pi^{\theta_N}$
\begin{algorithmic}[1]
\FOR {iteration n=0,1,...N}
\STATE Sample a set of trajectories $\{\tau\} \sim \pi^{\theta_n}$ consisting of tuples $(s, a, r, c_1, .., c_K)$
\STATE Compute advantage function $A^{\pi_n}$ and cost advantage functions $A^{\pi_n}_{C_{1,..K}}$ using GAE
\STATE Update constraint thresholds with (\ref{eq:adaptive_constraint_threshold})
\STATE Update policy parameters $\theta$ with (\ref{eq:ipo_problem_approx})
\STATE Update value function parameters $\phi$ and multi-head cost value function parameters $\psi$
\ENDFOR
\RETURN policy parameters $\theta=\theta_{N}$
\end{algorithmic}
\end{algorithm}

\section{Applications on Legged Robot Locomotion}
\label{sect:legged_robot}

We applied the proposed learning framework, consisting of both rewards and constraints, to train locomotion controllers for legged robots. Designing robust controllers for legged robots is challenging due to the underactuated nature and the complexity of the hardware. Recently, various works \cite{lee2020blind, miki2022perception, ji2022concurrent, choi2023deformable} used model-free deep reinforcement learning to design controllers and demonstrated impressive control performance in perspective of both robot's speed and robustness. However, the reward-only framework, that they utilized, required a lot of effort in reward engineering (more than ten reward terms were designed and the reward coefficients for each of them were tuned laboriously) to create a natural motion. In this work, we train high-performance locomotion controllers with significantly fewer reward terms to engineer by leveraging more intuitive and generalizable constraints.

\subsection{Overview}
Our objective is to devise a blind locomotion controller for legged robots, enabling them to navigate demanding terrains, as exemplified by the work of Lee et al. \cite{lee2020blind}. Blind locomotion controllers solely depend on proprioceptive sensor data (e.g., joint positions, body orientation) to adjust base motions and foot placements. The controller is given a velocity command, which consists of the desired forward velocity [m/s], lateral velocity [m/s], and turning rate [rad/s], and should track them as closely as possible even under unexpected terrain variations. 

Our training pipeline is built based on the current state-of-the-art works \cite{lee2020blind, miki2022perception} that leverage teacher-student learning in the physics simulation and transfer the policy to the real world in zero-shot. In these works, the teacher policy is first trained with model-free reinforcement learning using proprioceptive sensor information and privileged information only available in the simulation. The student policy is subsequently trained to mimic the behaviors of the teacher policy and forecast compressed privileged information based on the history of proprioceptive sensor data. The student network is trained by supervised learning. We use the proposed learning framework consisting of both rewards and constraints for training the teacher policy. Unlike the previous works that represented hardware constraints and motion styles as rewards, we directly incorporate them as constraints into our problem formulation. We found that only three reward terms are sufficient to obtain a natural gait policy that performs on par with the state-of-the-art policy.

Broad interaction data is crucial to make the controller robust against various rough terrains in the real world. To this end, parameterized terrains with various geometries are procedurally generated during training. The terrains are generated in a way that is not too difficult for the current policy to learn \cite{xie2020allsteps}. Domain randomization \cite{peng2018domainrandom} is done during training for some possible physical properties that can cause a sim-to-real gap. The overall learning framework is summarized in Figure \ref{fig:learning_framework}.

\subsection{Teacher policy}
We formulate the control problem as a Constrained Markov Decision Process (CMDP). In this subsection, we provide the core elements of the CMDP for teacher training consisting of state space, action space, reward function, and cost functions. 

The state space is defined as $s_t := <o_t, p_t>$ where $o_t$ is the observation directly accessible in the robot from various sensors and $p_t$ is the privileged information only available in the physics simulation. $o_t$ contains the given velocity command, body orientation, body angular velocity, current joint positions and velocities, joint position errors and velocities measured at (-0.02s, -0.04s, -0.06s), action histories at (-0.01s, -0.02s), and central phases for each leg. Central phases, defined as cosine and sine pairs $(cos(\Phi(t)), sin(\Phi(t)))$ of internal clocks $\Phi(t)=2\pi ft + \Phi_0$ of the robot, are used to depict desired gait patterns by altering gait frequency $f$ and initial phase shift $\Phi_0$. $p_t$ consists of information that is not directly available from the robot's measurements but can be useful for control and be approximately predicted from measurement histories. Specifically, it includes terrain friction coefficient, body height with respect to the ground, body linear velocity, body contact state, foot contact impulse, foot airtime, foot clearance with respect to the ground, and terrain profiles. Terrain profiles are given to the robot in the form of circular height scans near each foot similar to Lee et al. \cite{lee2020blind}.

The action space is defined as joint PD targets which are converted to joint torques by a PD controller module at a higher control frequency (4000Hz). For stable training, we follow the technique used by Ji et al. \cite{ji2022concurrent} where the action space is parameterized with nominal joint positions and an action scaling factor that is set constant during training. Specifically, the desired P target is computed as $q^{des}_t = q^{nominal} + \sigma_a \cdot a_t$, where $q^{nominal}$ is the nominal joint configuration, $\sigma_a$ is the action scaling factor, and $a_t$ is the neural network policy output. D target is set to zero. All the robots that we used in the experiment have an action space dimension of 12 and $\sigma_a$ is set to 0.4.

\begin{table}[!ht]
\vspace*{-0.15cm}
\caption{Reward Functions}
\vspace*{-0.25cm}
\label{table:reward}
\begin{center}
\begin{tabular}{|p{0.97cm}|p{0.97cm}|p{0.97cm}|p{0.97cm}|p{0.97cm}|p{1.03cm}|}
\hline
\multicolumn{2}{|c|}{\textbf{Reward}} & \multicolumn{4}{c|}{\textbf{Expression}} \\ \hline
\multicolumn{6}{|l|}{Command tracking} \\ \hline
\multicolumn{6}{|l|}{
$r_c=-k_c (\vert\vert cmd_{v_{xy}}-V_{xy}\vert\vert^2 + (cmd_{w_{z}}-w_{z})^2) $
} \\ \hline
\multicolumn{2}{|c|}{Joint torque} & \multicolumn{4}{l|}{$r_{\tau} = -k_{\tau}\vert\vert \tau\vert\vert ^2$} \\ \hline
\multicolumn{6}{|l|}{Action smoothness} \\ \hline
\multicolumn{6}{|l|}{
$r_s = -k_s (\vert\vert q^{des}_t-q^{des}_{t-1}\vert\vert ^2 + \vert\vert q^{des}_t-2q^{des}_{t-1}+q^{des}_{t-2}\vert\vert ^2)$
} \\ \hline
\end{tabular}
\end{center}
\vspace*{-0.25cm}
\end{table}

Rewards are composed of three terms: command tracking, joint torque, and action smoothness. Command tracking reward and joint torque reward guide the robot to track the given velocity command with minimal torque usage. Action smoothness reward is additionally used to regulate the neural network policy to have a smooth output surface, which is crucial for preventing motor vibration in a real-world deployment. The detailed reward equations are in Table \ref{table:reward}. The final reward given to the robot is $r = r_c + r_{\tau} + r_s$, where $k_c, k_{\tau}, k_s$ are reward coefficients representing relative weights between different reward terms. To make the training framework more generalizable across different robot platforms, we used the linear relationship between the energy usage (i.e., torque usage $\vert\vert \tau \vert\vert^2$) and the robot's mass. To this end, we parameterized the torque reward coefficient $k_{\tau}$ of a robot with mass $m$ as $k_{\tau} = \hat{k}_{\tau} \cdot \bar{m} / m = (s_{\tau} \cdot \bar{k}_{\tau}) \cdot \bar{m} / m$, where $\hat{k}_{\tau}$ is the torque reward coefficient before mass compensation, $s_{\tau}$ is the scaling factor, $\bar{k}_{\tau}$ and $\bar{m}$ each are the torque reward coefficient and the mass of the reference robot. We first find an adequate torque reward coefficient for the reference robot $\bar{k}_{\tau}$ by trial and error. For the reference robot, $s_{\tau}=1$ and $m=\bar{m}$. When training for a different robot, we initialize the torque reward coefficient to $k_{\tau}=\bar{k}_{\tau} \cdot \bar{m} / m$, where $s_{\tau}=1$, and modify $s_{\tau}$ as we observe to their resulting motions. In our work, a quadruped robot Raibo \cite{choi2023deformable} is the reference robot.

Cost functions are defined based on their constraint type. If the constraint should be satisfied at every time step or is easier to be represented via probability, we formulate it as a probabilistic constraint (\ref{eq:probabilistic_constraint}) and use an indicator function for the cost function (\ref{eq:probabilistic_constraint_cost_func}). If not, the constraint is formulated as an average constraint (\ref{eq:average_constraint}) by defining the cost function as the corresponding physical variable (\ref{eq:average_constraint_cost_func}).

Below terms are defined as probabilistic constraints.
\begin{itemize}
    \item \emph{Joint position ($c_{jp}$)}: Limit the upper and lower bounds of each joint angle to be in a feasible or desirable operational space. The cost function is an indicator function giving $0$ if the current joint angle is in the desirable range, and $1 / n_{joints}$ if not.
    \item \emph{Joint velocity ($c_{jv}$)}: Limit the upper and lower bounds of joint velocities according to the specifications of joint motors. The cost function is an indicator function giving $0$ if the current joint velocity is in the desirable range, and $1 / n_{joints}$ if not.
    \item \emph{Joint torque ($c_{jt}$)}: Limit the upper and lower bounds of joint torques according to the specifications of joint motors. The cost function is an indicator function giving $0$ if the current joint torque is in the desirable range, and $1 / n_{joints}$ if not.
    \item \emph{Body contact ($c_{bc}$)}: Avoid contact between terrains and the rest of the body except the feet. The cost function is an indicator function giving $0$ if the detected contact is between the terrain and the feet, and $1$ if not (e.g., self-collision, contact between the terrain and trunk/hip/thigh).
    \item \emph{Center of Mass (COM) frame ($c_{com}$)}: Limit the COM height to a reasonable range from the ground and the COM frame orientation to a desirable range with regard to gravity. The cost function is an indicator function giving $0$ if the COM frame is in the desirable range, and $1$ if not (e.g., COM height too close to the ground, COM orientation too tilted with respect to the gravity vector).
    \item \emph{Gait pattern ($c_{gp}$)}: Approximately match the predefined foot contact timing designed by the control engineer (e.g., trot, bound). The contact timing is defined based on the central phases for each leg by setting appropriate $f$ and $\Phi_0$, where the foot is in a swing phase if $sin(\Phi(t)) < 0$ and a stance phase if not. The cost function is an indicator function giving $0$ if the current foot contact state matches the desired state, and $1 / n_{legs}$ if not.
\end{itemize}
The desirable range $\textbf{S}$ defining the indicator cost functions of \textit{joint angle}, \textit{joint velocity}, and \textit{joint torque} constraints can be retrieved automatically from the robot description file (e.g., URDF files) if available, or manually determined by the engineer based on the desired motion to obtain or prior knowledge of the system to control. The constraint limits for the above constraints, except \textit{gait pattern} constraint $c_{gp}$, are set to a fairly small value to restrict the policy from entering the undesirable region.

Below terms are defined as average constraints.
\begin{itemize}
    \item \emph{Orthogonal velocity ($c_{ov}$)}: Limit the body velocity in a direction not given in the velocity command (i.e., $v_{ortho} = (v_z, w_x, w_y)$) to avoid noisy body motions. The cost function is defined as $c_{ov} = \vert \vert v_{ortho} \vert \vert_1 / 3$.
    \item \emph{Contact velocity ($c_{cv}$)}: Limit the velocity at contact points to prevent slippage. The cost function is the average of all contact point velocities $c_{cv} = \sum_{\forall contact} \vert \vert (v_x, v_y) \vert \vert_2 / N_{contact}$
    \item \emph{Foot clearance ($c_{fc}$)}: Limit the foot clearance to be above a desired value. Foot clearance is defined as the foot height from the ground at the peak of a swing phase (i.e., $sin(\Phi_t) \approx -1$). Foot clearance for each foot is updated at every leg phase period. The cost function is defined as $c_{fc} = -\sum_{\forall i \in \{1,..n_{leg}\}} D_{leg}^i / n_{leg}$ where $D_{leg}^i$ is the foot clearance of leg $i$. Minus is multiplied because the constraints are defined with upper bounds as in Eq. \ref{eq:constrained_rl_problem}.
    \item \emph{Foot height limit ($c_{fh}$)}: Limit foot height with respect to the ground to prevent raising them too high. The cost function is defined as $c_{fh} = max(d_{leg}^1, d_{leg}^2, ..d_{leg}^{n_{leg}})$ where $d_{leg}^i$ is the current foot height of leg $i$ with respect to the terrain profile near each foot.
\end{itemize}
The average constraints have the advantage of allowing the constraint thresholds to be defined intuitively because they directly correspond to physical quantities with known units. (e.g., velocity [\si{m/s}], height [\si{m}]). For example, if the control engineer wants the robot to raise the foot higher than $0.1\si{m}$, then the threshold of \textit{foot clearance} constraint just needs to be set to $-0.1$. Thresholds for the \textit{orthogonal velocity} and \textit{contact velocity} constraints are set to small values based on the logs obtained during the training.

Additionally, we constrain the body motion to be generated symmetrically by utilizing the symmetric loss proposed by Yu et al. \cite{yu2018symmetric}. The symmetric constraint $c_{sym}$ is defined as follows:
\begin{equation}
L_\mathrm{sym} := \underset{s \sim d^{\pi}}{\mathbb{E}}\left[\lVert\mu_\theta(s) - \Psi_a(\mu_\theta(\Psi_s(s)))\rVert_1\right] \leq d_\mathrm{sym},
\end{equation}
where $\Psi_a$ and $\Psi_s$ are functions that mirror the action and state with respect to the XZ plane of the base frame respectively, and $\mu_\theta$ is the mean of the Gaussian policy $\pi$ parameterized with neural network parameters $\theta$. The symmetric constraint is handled in the same manner as the average constraint by using the reparameterization trick \cite{kingma2013vae}.

To summarize, a total of 11 constraints (6 probabilistic constraints and 5 average constraints) are defined and used to train a control policy for legged robots. To the best of our knowledge, this is the greatest number of constraints ever used in the constrained reinforcement learning literature. Some of the constraints may not be critical during the optimization depending on the robot and the environment. For example, for some robots, the \textit{joint velocity} constraint may have less of an impact because the desirable range was chosen with enough margin, and so the initial policy already satisfied the desired constraint threshold. However, constraints do not need to be switched on and off according to their usage because of the property of logarithmic barrier functions. Concretely, already satisfied constraints have no effect on policy optimization due to a near-zero gradient of the barrier function. However, the barrier function applies a penalty when the policy tries to exit the desired region during exploration via a steep gradient.

The teacher policy is treated as a Gaussian policy, with the neural network outputs and state-independent trainable parameters corresponding to the distribution's mean and standard deviation. The teacher policy network is constructed with two Multi-Layer Perceptron (MLP) blocks: the MLP encoder and the MLP actor. The MLP encoder takes the privileged information $p_t$ as input and encodes it to a latent representation $l_t$. $l_t$ is then concatenated with the observation $o_t$ and passed to the MLP actor which outputs the action $a_t$.

\subsection{Student policy}
The student policy is trained to imitate the teacher policy's behavior. As the student policy is to be deployed in the real world in a zero-shot manner, it should only leverage information directly available from the robot's sensor and cannot utilize the privileged information $p_t$ used for the teacher policy. Thus, the student policy is trained to imitate the teacher policy's action $a_t$ while predicting the encoded privileged representation $l_t$ from observation histories $\{o_0, o_1, ..o_t\}$.

For this purpose, the student policy network is constructed with a Gated Recurrent Units (GRU) \cite{chung2014gru} encoder and an MLP actor. The GRU encoder takes in the observation $o_t$ and the internal hidden state of the unit and predicts the encoded privileged information $\bar{l}_t$. $\bar{l}_t$ is then concatenated with $o_t$ and passed to the MLP actor to output the action $\bar{a}_t$. The network is trained end-to-end with the loss function $L_{student}(\theta) = \vert\vert \bar{a}_t(\theta) - a_t \vert\vert_2^2 + \vert\vert \bar{l}_t(\theta) - l_t \vert\vert_2^2$, where $\theta$ is the neural network parameters of the student policy, to imitate both the teacher's action and the latent privileged representation.

\subsection{Terrain curriculum}
Similar to the previous works \cite{lee2020blind, miki2022perception, ji2022concurrent, choi2023deformable}, broad parameterized terrains are generated in the simulation by sampling the corresponding terrain parameters in a desired range. Five different types of terrain (i.e., hills, discrete hills, steps, inclining steps, and stairs) are sampled in an equal proportion and each type is modeled with two different terrain parameters similar to previous works \cite{lee2020blind, miki2022perception}.

During teacher policy training, an adaptive terrain curriculum is held to procedurally generate terrains that can give rich training signals to the network. The method we use is similar to the fixed-order curriculum proposed by Xie et al. \cite{xie2020allsteps}. The total ranges for each terrain parameter are divided into $N_{stage}$ and at each iteration, terrains are sampled uniformly between $0$ and $N_{max}^i$ stage where $N_{max}^i \ (\forall i \ N_{max}^i \in \{0,..N_{stage}\})$ is the maximum available sampling stage for $i$th terrain type. $N_{max}^i$ is set to $0$ at the beginning of the training and procedurally grows whenever the policy's traversability score at the $N_{max}^i$ stage exceeds the defined threshold. The traversability score is defined as Lee et al. \cite{lee2020blind} where the value is between 0 and 1. There are more sophisticated terrain curriculum methods available, such as the particle-filter approach \cite{lee2020blind}. In our experience, however, the strategy was still able to prevent catastrophic forgetting with essentially less related hyperparameter engineering, while still enabling strong policy training with effective terrain exploration.

The terrain curriculum is not employed during student policy training, but rather uniformly sampled in the terrain range for which the teacher policy is trained.

\subsection{Domain randomization}
For the zero-shot sim-to-real transfer of the trained control policy, critical parameters of the robot are randomized during the training. Specifically, observation noise, motor frictions, PD controller gains, foot positions and collision geometry, ground friction, and control latency are randomized in the predefined distribution at the start of the episode or at every time step.

\section{Experimental Results}\label{sect:experiment_results}

\begin{table*}[!ht]
    \centering
    \caption{Constraint satisfaction results}
    \begin{tabular}{|c|l|c|c|c|c|c|c|c|c|c|c|c|}
    \hline
    \multicolumn{2}{|c|}{} & $c_{jp}$ & $c_{jv}$ & $c_{jt}$ & $c_{bc}$ & $c_{com}$ & $c_{gp}$ & $c_{ov}$ [\si{m/s}] & $c_{cv}$ [\si{m/s}] & $c_{fc}$ [\si{m}] & $c_{fh}$ [\si{m}] & $c_{sym}$ \\
    \hline & Limit & 0.025 & 0.025 & 0.025 & 0.025 & 0.025 & 0.25 & 0.35 & - & - & - & 0.1 \\
    \hline
    \hline \multirow{7}{*}{Sec. \ref{sec:experiment:evaluation}} & Raibo & 0.015 & 0.004 & 0.001 & 0.002 & 0.009 & 0.16 & 0.32 & 0.15 [0.2] & -0.11 [-0.07] & 0.09 [0.11] & 0.07 \\
    \cline{2-13} & Hound & 0.014 & 0.020 & 0.000 & 0.003 & 0.002 & 0.17 & 0.30 & 0.13 [0.2] & -0.09 [-0.07] & 0.07 [0.11] & 0.07 \\
    \cline{2-13} & Go1 & 0.012 & 0.007 & 0.000 & 0.008 & 0.001 & 0.16 & 0.33 & 0.10 [0.2] & -0.08 [-0.05] & 0.06 [0.09] & 0.07 \\
    \cline{2-13} & Mini-cheetah & 0.015 & 0.000 & 0.001 & 0.002 & 0.001 & 0.13 & 0.32 & 0.10 [0.2] & -0.08 [-0.05] & 0.06 [0.09] & 0.06 \\
    \cline{2-13} & Anymal B & 0.016 & 0.019 & 0.000 & 0.002 & 0.005 & 0.16 & 0.31 & 0.13 [0.2] & -0.09 [-0.07] & 0.07 [0.11] & 0.07 \\
    \cline{2-13} & Anymal C & 0.019 & 0.021 & 0.002 & 0.004 & 0.005 & 0.16 & 0.30 & 0.11 [0.2] & -0.08 [-0.07] & 0.06 [0.11] & 0.07 \\
    \cline{2-13} & Atlas & 0.011 & 0.019 & 0.004 & 0.008 & 0.008 & 0.12 & 0.29 & 0.35 [0.5] & -0.17 [-0.15] & 0.12 [0.2] & 0.08 \\
    \hline \multirow{2}{*}{Sec. \ref{sec:experiment:analyzation}} & Torque (x2) & 0.017 & 0.005 & 0.001 & 0.001 & 0.008 & 0.15 & 0.33 & 0.16 [0.2] & -0.13 [-0.07] & 0.10 [0.11] & 0.08 \\
    \cline{2-13} & Smoothness (x2) & 0.013 & 0.005 & 0.001 & 0.001 & 0.009 & 0.14 & 0.32 & 0.15 [0.2] & -0.13 [-0.07] & 0.10 [0.11] & 0.07 \\
    \hline \multirow{5}{*}{Sec. \ref{sec:experiment:ablation}} & P3O \cite{gangapurwala2020gcpo, zhang2022p3o} & \textbf{0.025} & 0.005 & 0.000 & 0.002 & 0.016 & 0.24 & \textbf{0.35} & \textbf{0.2} [0.2] & -0.08 [-0.07] & 0.09 [0.11] & \textbf{0.1} \\
    \cline{2-13} & $t=100$, $\alpha=0.002$ & 0.015 & 0.004 & 0.002 & 0.003 & 0.009 & 0.17 & 0.32 & 0.15 [0.2] & -0.11 [-0.07] & 0.09 [0.11] & 0.07 \\
    \cline{2-13} & $t=100$, $\alpha=0.2$ & 0.015 & 0.004 & 0.002 & 0.002 & 0.011 & 0.18 & \textbf{0.44} & 0.17 [0.2] & -0.11 [-0.07] & 0.09 [0.11] & 0.07 \\
    \cline{2-13} & $t=10$, $\alpha=0.02$ & 0.004 & 0.001 & 0.000 & 0.000 & 0.002 & 0.11 & 0.24 & 0.07 [0.2] & -0.14 [-0.07] & 0.07 [0.11] & 0.05 \\
    \cline{2-13} & $t=1000$, $\alpha=0.02$ & 0.023 & 0.010 & 0.002 & 0.002 & 0.015 & 0.22 & \textbf{0.45} & \textbf{0.2} [0.2] & -0.09 [-0.07] & 0.09 [0.11] & 0.09 \\
    \hline
    \end{tabular}
    \begin{tablenotes}[flushleft] \footnotesize
     \item $^*$Results are after training with 1.6 billion time steps. Constraint thresholds are reported in the top row or inside the bracket. Bolded texts indicate violated or barely satisfied constraints.
    \end{tablenotes}
    \label{tab:contraint_satisfaction}
    \vspace*{-0.25cm}
\end{table*}

We conducted several experiments with the task of legged robot locomotion to answer the questions below regarding a learning framework that incorporates both rewards and constraints:
\begin{itemize}
    \item Can a controller trained with the proposed framework attain a similar robust control performance as prior works developed with the reward-only approach?
    \item Can the proposed framework be more generalizable across different robot platforms?
    \item Can the proposed framework make the engineering process of generating desired motions more straightforward and intuitive?
\end{itemize}

The rest of this section is structured as follows. After providing some implementation details in Sec. \ref{sec:experiment:implementation}, the results of training locomotion controllers for various legged robots are shown in Sec. \ref{sec:experiment:evaluation}. Some crucial characteristics
required for the usage of the learning framework are analyzed in Sec. \ref{sec:experiment:analyzation}. In Sec. \ref{sec:experiment:comparison}, the training performance is compared with the reward-only framework. Lastly, ablation studies are conducted in Sec. \ref{sec:experiment:ablation} for an element-wise in-depth study of the proposed framework.

\subsection{Implementation detail}\label{sec:experiment:implementation}
The proposed learning framework was implemented with Pytorch \cite{paszke2019pytorch} and publicly available constrained reinforcement learning algorithm implementations \cite{kimsaferlalgo2023}. To train locomotion controllers for legged robots in the physics simulator, the simulator should be capable of producing a large volume of realistic interaction data in a highly parallel and rapid manner. For this purpose, RaiSim \cite{hwangbo2018raisim} physics simulator was utilized. For training, we used AMD Ryzen9 5950X and a single NVIDIA GeForce RTX 3070. It took about 24 hours to train the teacher policy and 8 hours to train the student policy. 

The locomotion control policy runs at 100Hz. We deployed the trained controller on two real-world quadruped robots: Raibo \cite{choi2023deformable} and Mini-cheetah \cite{katz2019minicheetah}. For the deployment, the student policy network was reimplemented with the Eigen library in C++ and forward-passed in the Central Processing Unit (CPU). We recommend readers check the original paper \cite{choi2023deformable, katz2019minicheetah} for more details about the robot hardware.

We analyzed the performance of the proposed learning framework in various aspects, such as algorithm computation time and obtained rewards. The Raibo robot is used to produce the majority of simulation results; if another kind is used, it is noted.

\subsection{Framework Evaluation} \label{sec:experiment:evaluation}
\begin{figure*}[t!]
\centering
\includegraphics[width=\linewidth]{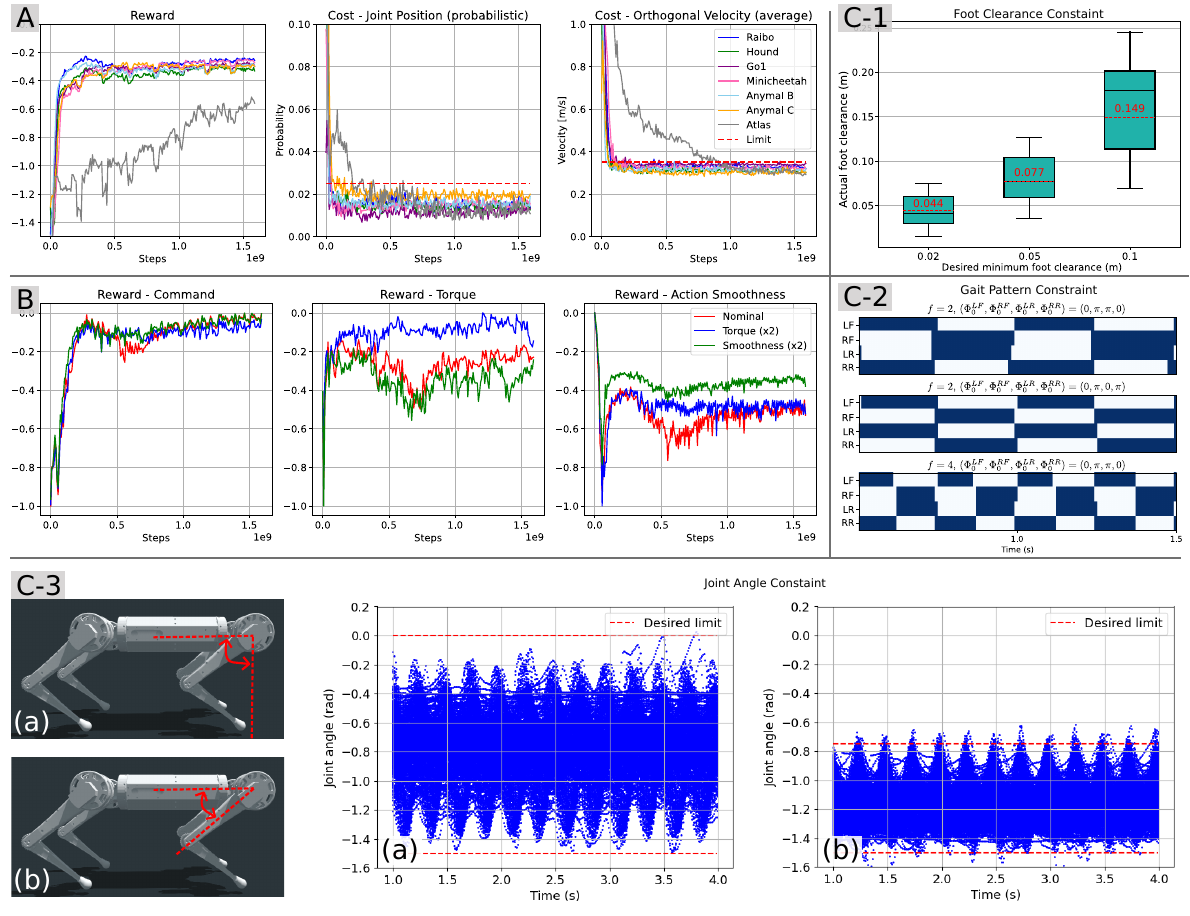}
\caption{Evaluation and analysis results of the proposed framework. (A) shows the total reward (left) and two costs (middle, right) of multiple robots. (B) shows each reward component's value to check the \textit{objective sensitivity} property. Every reward has been standardized so that it ranges from -1 to 0. (C) shows the results to verify the \textit{constraint satisfaction} property. Specifically, foot clearance (C-1), gait pattern (i.e., trot, pace, fast trot) (C-2), and joint angle (C-3) constraints were modified and tested in the simulation. The resulting foot clearances, foot contact patterns, and joint angle trajectories are plotted.}
\label{fig:framework_eval_and_analysis_sim}
\vspace*{-0.25cm}
\end{figure*}

\begin{figure*}[t!]
\centering
\includegraphics[width=\linewidth]{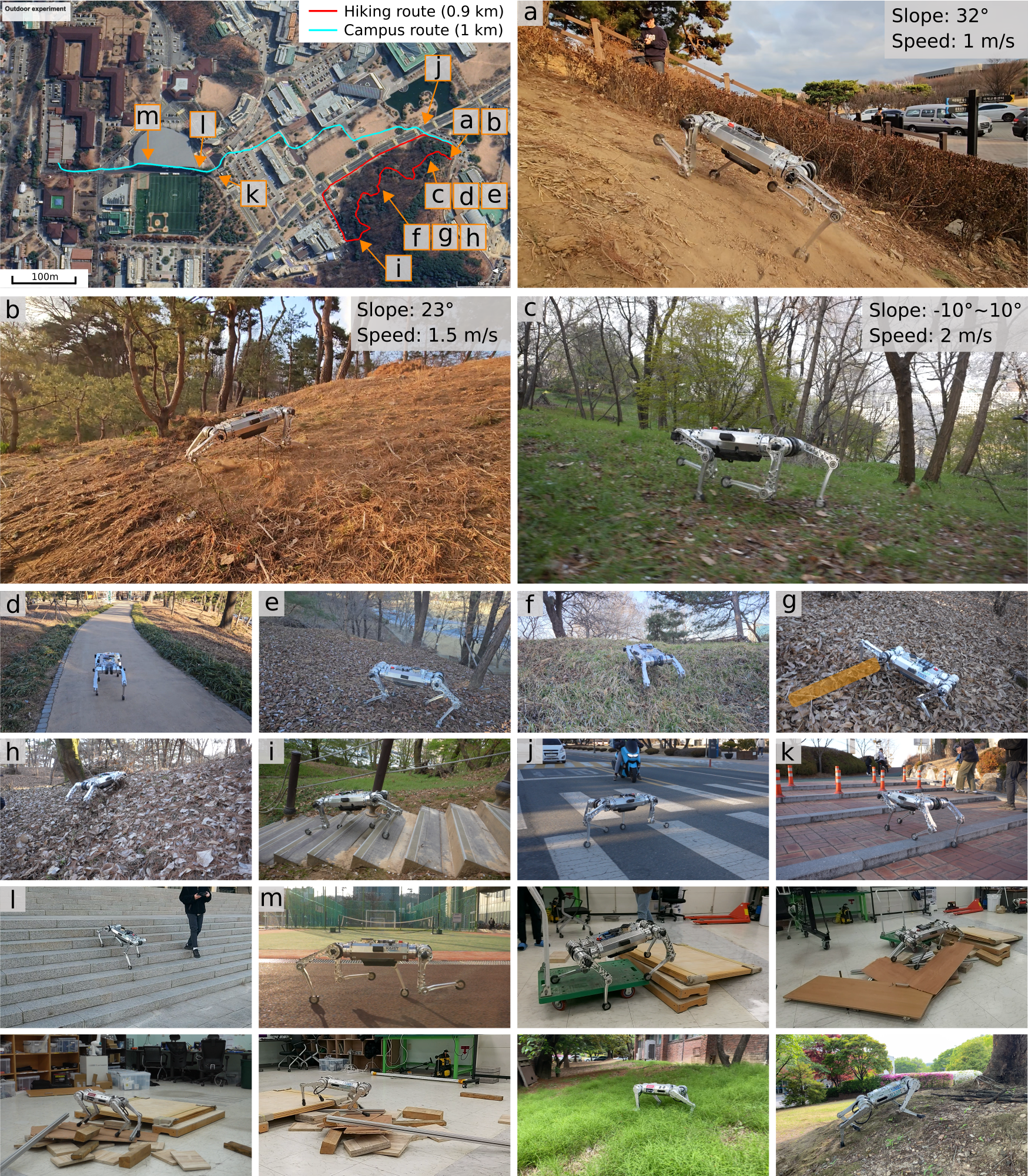}
\caption{Real-world experiments with two quadruped robots: Raibo and Mini-cheetah. The robustness of the controllers was extensively tested in various harsh terrains both indoor and outdoor. Outdoor experiments were conducted on two long paths: a hiking route and a campus route. The robot's speed is calculated using the video's recorded time stamp and the measured length of the traversal path. The slope of the environment is measured using a digital angle gauge.}
\label{fig:hardware_experiment}
\vspace*{-0.25cm}
\end{figure*}

\begin{figure*}[t!]
\centering
\includegraphics[width=\linewidth]{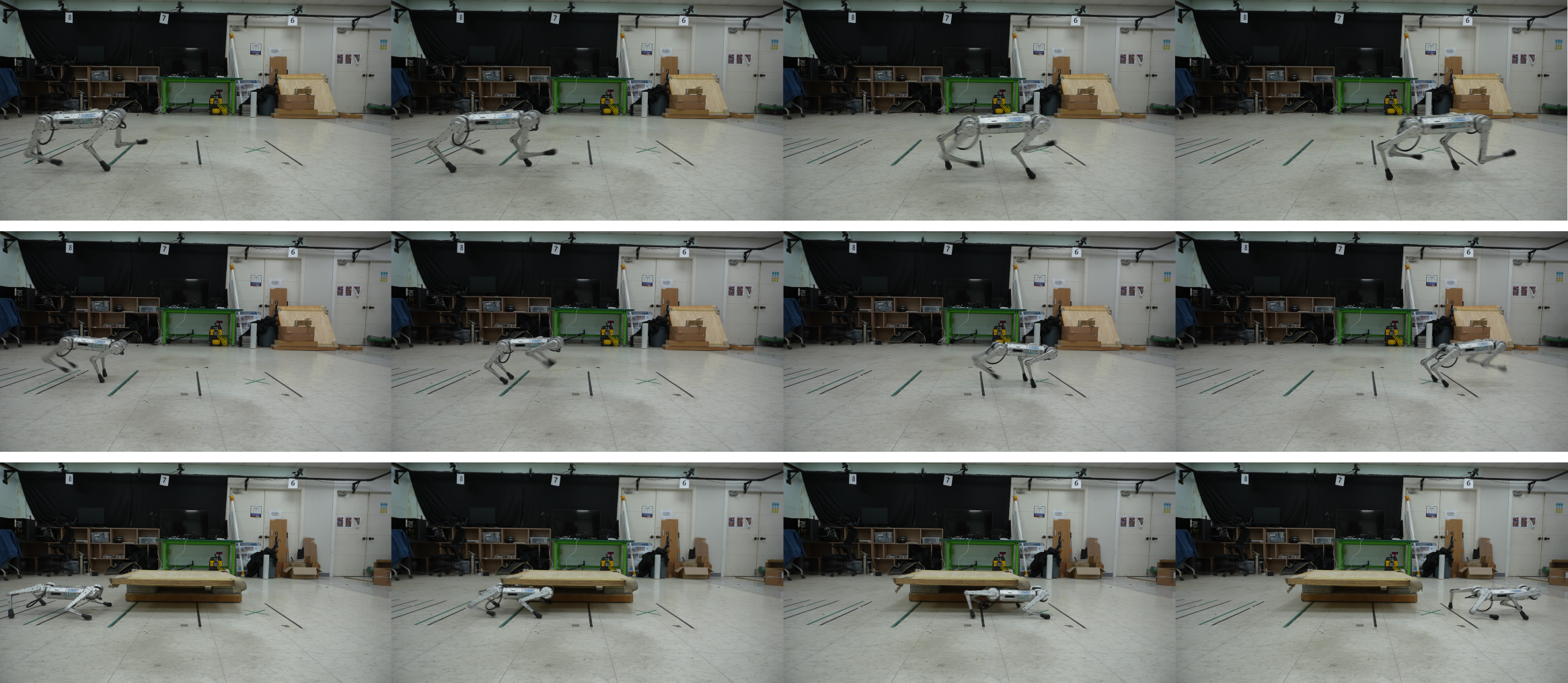}
\caption{Real-world experiments of generating multiple motions by varying constraints. A robot walking with a trotting gait (top) was changed to locomote with a bounding gait (middle) or crawl beneath an overhanging obstacle (bottom) by modifying the gait pattern constraint or COM height with joint angle constraint.}
\label{fig:framework_eval_and_analysis_real}
\vspace*{-0.25cm}
\end{figure*}

The proposed learning framework was used to train locomotion controllers for diverse legged robots in the physics simulator: Raibo \cite{choi2023deformable}, Mini-cheetah \cite{katz2019minicheetah}, Hound \cite{shin2022hound}, Anymal B, Anymal C \cite{hutter2016anymal}, Unitree Go1, and Atlas (Figure \ref{fig:multiple_robot}). Robots used for training vary in both morphology and physical properties. In particular, two robot types (quadruped and bipedal) are studied, with mass ranging from 11\si{kg} to 135\si{kg}, and robot size and leg length ranging from a 0.3\si{m} tall small-scale dog with a 0.4\si{m} leg length to a 1.9\si{m} tall person with a 0.9\si{m} leg length. Furthermore, suitable PD gains for each robot varied substantially depending on the robot's motor specification and inertia, and the hind legs design of quadruped robots (Anymal B and Anymal C) differed depending on whether they are bent forward or backward.

\begin{table}[!t]
    \centering
    \caption{Reward coefficients for several legged robots}
    \begin{tabular}{|c|c|c|c|}
    \hline
        & $k_c$ & $\hat{k}_{\tau}$ & $k_s$ \\
    \hline Raibo & 10. & 0.005 & 0.06 \\
    \hline Hound & 10. & 0.0025 & 0.06 \\
    \hline Go1 & 10. & 0.025 & 0.06 \\
    \hline Mini-cheetah & 10. & 0.03 & 0.06 \\
    \hline Anymal B & 10. & 0.005 & 0.06 \\
    \hline Anymal C & 10. & 0.003 & 0.06 \\
    \hline Atlas & 10. & 0.001 & 0.06 \\
    \hline
    \end{tabular}
    \label{tab:general_robot_reward_coeff}
    \vspace*{-0.25cm}
\end{table}

Regardless of these variations, robust locomotion controllers for individual robots could be trained successfully using our framework with minimal parameter modifications. As shown in Figure \ref{fig:framework_eval_and_analysis_sim}-A and Table \ref{tab:contraint_satisfaction}, policies gradually learned to maximize the designed rewards while satisfying all of the constraints. Scaling factor $s_{\tau}$, which defines the torque reward coefficient $k_{\tau}$, was only adjusted among the reward coefficients for different robots (Table \ref{tab:general_robot_reward_coeff}). This is reasonable because energy usage is a complex function that cannot be solely modeled by mass; it is also influenced by factors like link inertia and actuator characteristics.

Constraint limits and desirable region $\textbf{S}$ for some probabilistic constraints were adjusted based on the robot's morphology and size. $\textbf{S}$ for the \textit{joint angle} constraints were set semantically the same for all quadruped robots. In the case of Atlas, a bipedal robot, the desirable joint angle ranges $\textbf{S}$ for the hip joints were manually set in the \textit{joint angle} constraint based on the related online video footage (e.g., the robot's locomotion videos) to prevent the robot from spreading its legs too far to the sides while walking. In the \textit{gait pattern} constraint, the size of the initial phase offset parameter $\Phi_0$ for quadrupedal robots was four (e.g., [0, pi, pi, 0]), while it was two (e.g., [0, pi]) for bipedal robots. This is necessary because each robot has a different number of legs. Thresholds for the \textit{COM}, \textit{foot clearance}, and \textit{foot height limit} constraints were adjusted based on the size of the robot. For instance, the 7cm foot clearance limit used for Anymal C is too large for a small-scale robot like Mini-cheetah. The \textit{contact velocity} constraint threshold was set the same for all quadruped robots but was slightly increased for Atlas. This was necessary because the foot's collision body of the quadruped robots was a sphere, whereas that of Atlas was a box. Desirable regions for \textit{joint velocity} and \textit{joint torque} constraints were set automatically from the robot description file (e.g., URDF files). Other constraint limits, desirable regions $\textbf{S}$, and reward functions were set exactly the same for all robots. Specific limits for each constraint are provided in Table \ref{tab:contraint_satisfaction}.

The trained controllers were deployed on two real-world quadruped robots available in the lab: Raibo and Mini-cheetah. The controllers were extensively tested in various harsh terrains both indoor and outdoor to verify the robust control performance as shown in the previous work \cite{lee2020blind}. The testing environments included slippery hills, steep slopes, deformable terrains, stairs, a pile of leaves with hidden impediments, a moving cart, unstably piled planks, and even more. No matter how difficult the terrains were, Raibo and Mini-cheetah were able to walk steadily by modifying base motions and foot placements based solely on proprioceptive sensor data, as illustrated in Figure \ref{fig:hardware_experiment}. Furthermore, Raibo could sprint over grassy terrain at 2m/s and climb slippery and steep slopes with an angle of around 32$^\circ$ at 1m/s which is roughly double the speed reported in the prior work \cite{lee2020blind}.

Real-world constraint satisfaction was also verified for those that can be directly observed by sensors (i.e., joint angle, joint velocity, joint torque). For verification, we collected approximately 25 minutes of real-world robot sensor data with the Raibo robot in the outdoor experiment routes shown in Figure \ref{fig:hardware_experiment}. \textit{Joint angle}, \textit{joint velocity}, and \textit{joint torque} constraints are all probabilistic constraints, and their cost values were 0.007, 0.001, and 0.002, respectively, indicating constraint satisfaction given that the thresholds were all 0.025.

In summary, locomotion controllers for various legged robots were trained with our framework that utilized only three rewards and generalizable constraints. A single locomotion learning framework could train control policies for N robots with a minimal amount of engineering. Specifically, only a single reward coefficient was adjusted among the three designed reward terms for robot transfer. If the robot's training is solely guided by a composition of abundant reward terms, the interdependency between them makes the additional engineering process difficult and cumbersome. More analysis about the reward-only framework is provided in Sec. \ref{sec:experiment:comparison}. There were some modifications for the constraints, but this can be considered a much more intuitive process for the control engineer compared to the reward coefficient tuning because the constraint bounds all have exact physical meanings. Our extensive real-world experiments show that dynamic and agile controllers trained in simulation can be successfully transferred to the real world in zero-shot and show highly robust performance in various harsh terrains.

\subsection{Framework Analyzation}\label{sec:experiment:analyzation}
The proposed learning framework leverages both rewards and constraints, where these two representations are used for different purposes. Rewards are representations of a value to be maximized and can be thought of as an objective function in numerical optimization problems. Constraints, on the other hand, are representations in which engineers express the desirable area that the converged policy should satisfy. To leverage both rewards and constraints for designing neural network controllers, the proposed framework should satisfy below two properties.
\begin{itemize}
    \item \textit{Objective sensitivity}: By varying the reward coefficients, which are comparable to altering the relative weights of various objective functions, the policy training should show responsive results. 
    \item \textit{Constraint satisfaction}: Regardless of the variations of objectives and constraints, the converged policy should satisfy defined constraints if there exists a policy that satisfies all constraints.
\end{itemize}

We conducted several experiments to verify these properties. First, we modified objectives by applying a constant scaling factor of two to coefficients of torque reward and smoothness reward, wherein the values are not bounded, and analyzed the policy training results. This is frequently done when engineers want to emphasize a specific term more when constructing the objective function, such as giving higher weights to the torque usage term to minimize them more. As shown in Figure \ref{fig:framework_eval_and_analysis_sim}-B, the policy's specific reward term grew as the corresponding coefficient increased. Additionally, the constraints were all satisfied regardless of the modifications of the reward signals (Table \ref{tab:contraint_satisfaction}). This is a promising result for the learning framework because the policy is trained to be responsive to the variation of the objective while satisfying the already existing constraints, which are both defined by the engineer's intent.

Second, we varied the constraints and saw aspects of the converged policies. Desirable regions $\textbf{S}$ and constraint limits were each modified for the probabilistic constraint and the average constraint. Mini-cheetah robot was used for this experiment. As shown in Figure \ref{fig:framework_eval_and_analysis_sim}-C, varying the constraints guided the policy to different behaviors that belong in the defined region. Specifically, the robot was trained in the simulation to have various foot clearance, walk in different gait patterns, and operate in a different joint angle range by changing the constraints. The effectiveness of using constraints was further analyzed by employing the controllers with several motions in real-world scenarios (Figure \ref{fig:framework_eval_and_analysis_real}). The robot was trained to crawl below an overhanging obstacle or utilize a bounding gait rather than a trotting gait. For the robot to crawl below an overhanging obstacle, the COM height range for the \textit{COM} constraint and the hip joint range for the \textit{joint angle} constraint were modified. For the robot to locomote with a bounding gait, the initial phase offset parameter $\Phi_0$ for the \textit{gait pattern} constraint was changed to represent them. The above two motions were generated from the exactly same parameter set that was used to train a walking controller with a trotting gait. 

\begin{table}[!ht]
\vspace*{-0.15cm}
\caption{Number of Reward Terms}
\vspace*{-0.25cm}
\label{table:reward_number}
\begin{center}
\begin{tabular}{|c|c|}
\hline
Lee et al. \cite{lee2020blind} & 6 \\ \hline
Miki et al. \cite{miki2022perception} & 9 \\ \hline
Kumar et al. \cite{kumar2021rma} & 10 \\ \hline
Ji et al. \cite{ji2022concurrent} & 11 \\ \hline
Choi et al. \cite{choi2023deformable} & 14 \\ \hline \hline
\textbf{Ours} & \textbf{3} \\ \hline
\end{tabular}
\end{center}
\vspace*{-0.25cm}
\end{table}

As a result, our framework satisfies two important properties, \textit{objective sensitivity} and \textit{constraint satisfaction}, required for training neural network controllers. Based on the \textit{constraint satisfaction} property, engineers can use constraints to restrict the area where the policy converges. Thus, they can generate desired motions efficiently by setting appropriate constraint parameters at their intent. Setting constraints is more straightforward than modifying reward coefficients since it directly restricts the search region with hyperparameters that correspond to physical variables with known units. A minimal number of reward terms, as demonstrated in Table \ref{table:reward_number}, can then be designed and fine-tuned based on the \textit{objective sensitivity} property. Check out the Appendix for a more detailed demonstration of our framework's engineering benefits based on the two properties mentioned above. It is illustrated using a simple toy example and a comprehensive comparison of the entire engineering scenario with the reward-only framework (Table \ref{table:engineering_scenario_comparison}).

\subsection{Comparison with the Reward-only Framework}\label{sec:experiment:comparison}
We compare the generalizability of our framework and the reward-only framework, which is dominantly used in previous works on legged robot locomotion \cite{lee2020blind, miki2022perception, ji2022concurrent, choi2023deformable}. The reward terms and hyperparameters used for the reward-only framework are based on the previous works by Ji et al.\cite{ji2022concurrent} and Choi et al. \cite{choi2023deformable}. There were a total of twelve reward terms and they were designed for task completion (i.e., command tracking), physical constraints' satisfaction (i.e., undesirable body contact, joint velocity, joint acceleration, joint torque, action smoothness), and style constraints' satisfaction (i.e., air time, foot slippage, foot clearance, nominal configuration, hip joint usage, orthogonal velocity). Adequate reward coefficients for each term were laboriously engineered for robust control performance on the Raibo robot while satisfying all physical constraints. TRPO was used for policy optimization.

\begin{figure}[t!]
\centering
\includegraphics[width=\linewidth]{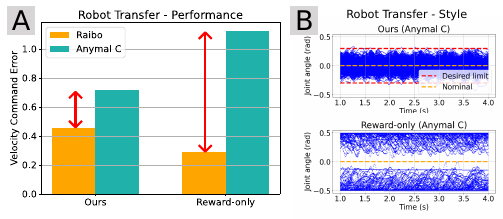}
\caption{Comparison results with the reward-only framework in terms of the robot generalizability. Each framework was engineered for robust control performance on the Raibo robot and was directly applied to the Anyaml C robot. (A) shows the locomotion performance according to the velocity command tracking error $\vert\vert cmd_{v_{xy}}-V_{xy}\vert\vert^2 + (cmd_{w_{z}}-w_{z})^2$. A smaller difference between Raibo and Anymal C indicates better robot generalizability. (B) shows the locomotion style of the Anymal C robot in terms of the hip joint angle trajectories. For the Raibo robot, both frameworks were engineered to utilize joint angle ranges near the nominal joint configuration. A smaller deviation from the nominal joint configuration indicates better robot generalizability.}
\label{fig:reward_only_comparison}
\vspace*{-0.25cm}
\end{figure}

The learning framework engineered for the Raibo robot was directly applied to the Anymal C robot, which is twice heavier and has a different leg morphology. None of the reward coefficients were modified for both methods, except the auto-adaptation of the torque reward coefficient $k_{\tau}$ based on the robot's mass. The scaling factor $s_{\tau}$ is set to one for both methods. As illustrated in Figure \ref{fig:reward_only_comparison}-A, Anymal C robot, trained with our framework, exhibited remarkable locomotion capabilities with comparable command tracking performance \footnote{Check out the Appendix for a more in-depth demonstration of the Raibo robot's velocity command tracking performance.}. On the other hand, the controller trained with the reward-only framework exhibited poor locomotion performance and significant degradation in its overall command tracking capability. Furthermore, the converged control policy for the Anymal C robot utilized joint angle ranges quite apart from the nominal configuration, resulting in poor locomotion style transfer, as shown in Figure \ref{fig:reward_only_comparison}-B. These circumstances were prevented for the Raibo robot using an extra reward term (i.e., the nominal configuration reward) which penalizes deviations from the nominal configuration. However, the corresponding reward signal was too small for the Anymal C robot and the reward coefficients should be adjusted accordingly. In our framework, the locomotion style was more transferable across robots because they were defined as constraints rather than soft penalty terms as used in the reward-only framework. 

\begin{figure}[t!]
\centering
\includegraphics[width=\linewidth]{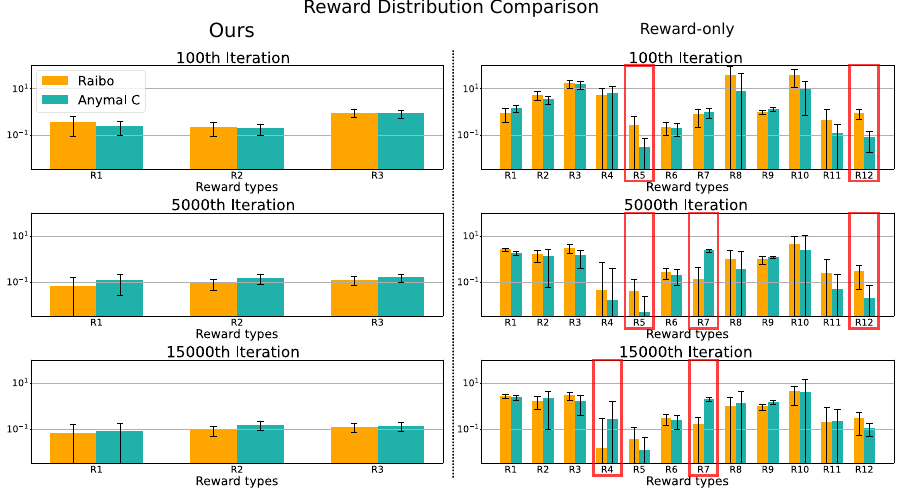}
\caption{Comparison of each reward component's distribution with the reward-only framework. Red blocks show reward types with highly different reward distributions.}
\label{fig:reward_only_comparison_reward}
\vspace*{-0.25cm}
\end{figure}

The poor generalizability of the reward-only framework is because the magnitudes of reward terms change depending on the robot. We plotted the mean and the standard deviation of each reward component for the two robots in Figure \ref{fig:reward_only_comparison_reward}. In the reward-only framework, the two robots showed very different reward distributions, thus resulting in different control behaviors. The changes in the reward distributions are due to the differences in the robot's physical properties and morphologies. Consequently, the reward coefficients should be adjusted, but it is not trivial due to the interdependency among each of the reward components and the reward signal distribution shift over the iteration number. This results in an additional cumbersome reward engineering process that involves a significant amount of trial and error to find the most suitable parameter set.

\begin{figure*}[t!]
\centering
\includegraphics[width=\linewidth]{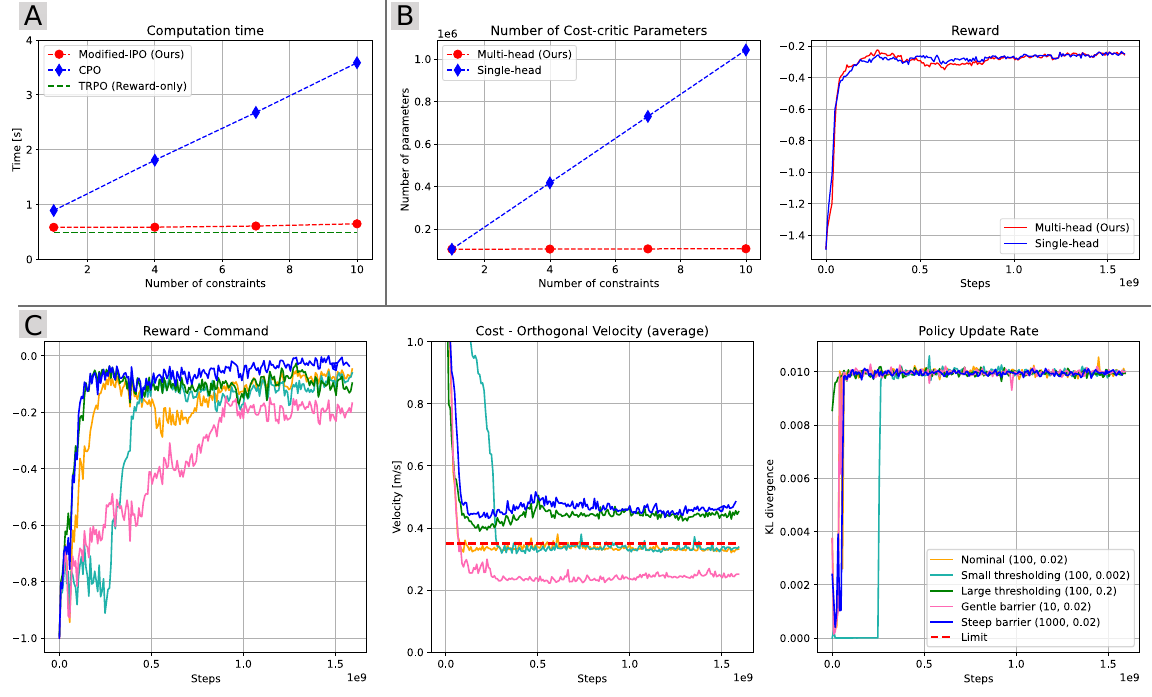}
\caption{Ablation study results of the proposed framework. (A) shows the computation time for different policy optimization algorithms. (B) shows the number of trainable parameters (left) and total rewards (right) for different cost-critic network designs. (C) shows the hyperparameter $t$ and $\alpha$ sensitivity in terms of the reward (left), cost (middle), and policy update rate (right).}
\label{fig:ablation}
\vspace*{-0.25cm}
\end{figure*}

\subsection{Ablation study}\label{sec:experiment:ablation}
We conducted several ablation studies to verify some core components of the proposed learning framework. Furthermore, the performance sensitivity to algorithm hyperparameters was analyzed.

\subsubsection{Algorithm}
Our policy optimization is based on the Interior-point Policy Optimization (IPO) \cite{liu2020ipo} which changes the constrained problem to an unconstrained problem using logarithmic barrier functions. We compared the optimization performance with two well-known policy-search algorithms for solving constrained reinforcement learning problems: \textit{Constrained Policy Optimization (CPO)} \cite{achiam2017cpo} and \textit{Penalized Proximal Policy Optimization (P3O)} \cite{zhang2022p3o}. CPO is an extension of Trust Region Policy Optimization (TRPO) \cite{schulman2015trpo} to the CMDP framework with linear approximations of the cost functions. They convert the primal problem to the dual problem to efficiently find the policy update direction. P3O, similar to IPO, is a first-order method that converts a constrained problem into an unconstrained problem for policy optimization. P3O applies a linear cost gradient penalty until the constraint is satisfied and then ignores the cost gradient within the constraint-satisfying area via the clamping function. Previously, a variant of P3O was used for training locomotion controllers for legged robots with successful real-world deployment \cite{gangapurwala2020gcpo}. Check the original papers \cite{achiam2017cpo, zhang2022p3o} for more detail.

As illustrated in Figure \ref{fig:ablation}-A, the computation time for the optimization step grows linearly with the number of constraints when employing CPO. This is because, the number of optimization variables in the dual problem, as well as the number of conjugate gradient methods to be applied, grows in proportion to the number of constraints. If there are one or two constraints, analytic solutions derived from the original paper can be utilized to reduce the computation time. However, we want our learning framework to be scalable enough to handle multiple constraints because it is intended to be used for training neural network controllers for complex robotic systems. IPO is a suitable optimization algorithm to utilize since it exhibits nearly no change in computation time with increasing the number of constraints, and it even has a similar computation time to the optimization step for the reward-only framework (TRPO). This is clear because our framework and the reward-only framework are completely identical when no constraints are given. The major extra computation that occurs when constraints are defined is the computation of the cost advantage functions and training of the cost value functions, which are parallelized using a multi-head version.

\begin{table}[h!]
\vspace*{-0.15cm}
\caption{Algorithm performance comparison}
\centering
\begin{tabular}{|ll|c|c|c|c|}
\hline
\multicolumn{2}{|l|}{}                               & Reward ($r$)          & Torque           & Smoothness      & Traversability        \\ \hline
\multicolumn{2}{|l|}{\multirow{2}{*}{\textbf{Ours}}} & \textbf{-0.242}  & \textbf{24.769}  & \textbf{0.093}  & \textbf{0.811}   \\  
\multicolumn{2}{|l|}{}                               &   ($\pm$ 0.007)  &   ($\pm$ 5.811)  &   ($\pm$ 0.002)  &  ($\pm$ 0.055)   \\ \hline
\multicolumn{2}{|l|}{P3O}           & -0.271           & 26.186           & 0.101           & 0.781            \\ 
\multicolumn{2}{|l|}{\cite{gangapurwala2020gcpo, zhang2022p3o}}                               & ($\pm$ 0.012)    &   ($\pm$ 6.116)  &   ($\pm$ 0.002)  & ($\pm$ 0.068)    \\ \hline
\end{tabular}
\label{table:p3o_comparison}
\end{table}

Because P3O is also a first-order method, it showed a similar computation time as our method and demonstrated that it can be scalable in terms of the number of constraints. However, the policy trained with IPO performed better than those trained with P3O (Table \ref{table:p3o_comparison}). The policy trained with IPO showed higher reward, smaller torque usage (i.e., smaller energy usage), better action smoothness (i.e., smaller motor vibration for real-world deployment), and a higher traversability score \cite{lee2020blind} (i.e., better ability to traverse diverse terrains at high speed). Furthermore, unlike the policy trained with IPO, which showed clear constraint satisfaction, the policy trained with P3O converged near the constraint threshold and exhibited constraint violation in some cases. (Table \ref{tab:contraint_satisfaction}). This is because IPO takes into account a smooth cost gradient within the constraint-satisfying region in the form of a logarithmic barrier function, whereas P3O completely disregards it. The smooth cost gradient within the constraint-satisfying region enables better policy guidance for high performance (e.g., the smooth gradient derived from the contact velocity constraint penalizes foot slippage even after satisfying the contact velocity constraint) and safety (e.g., the smooth gradient derived from the joint position constraint penalizes joint angle usage near the undesirable range).

\subsubsection{Multi-head cost value function}
We utilized the multi-head cost value function when computing the cost advantage functions using GAE \cite{schulman2015gae}. We compared the number of trainable parameters, computation time, and value prediction loss during training to the two approaches listed below.
\begin{itemize}
    \item \textit{Single-head}: Use independent neural networks with no shared parameters for each cost value function. The same neural network size with the multi-head method is applied for each cost value function.
    \item \textit{Single-head (Reduced)}: Use independent neural networks with no shared parameters for each cost value function. The reduced neural network size is applied for each cost value function to match the total number of trainable parameters with the multi-head method.
\end{itemize}

\begin{table}[h!]
\caption{Multi-head cost critic evaluation (\# constraints = 10)}
\centering
\begin{tabular}{|l|c|c|c|}
\hline
                                                                      & \# parameters & Computation time {[}\si{s}{]}                                   & Prediction loss                                         \\ \hline
\textbf{\begin{tabular}[c]{@{}l@{}}Multi-head \\ (Ours)\end{tabular}} & \textbf{0.1 M}         & \begin{tabular}[c]{@{}c@{}}\textbf{0.23} \\ (10.22 \%)\end{tabular} & \begin{tabular}[c]{@{}c@{}}0.468\\ ($\pm$ 0.120)\end{tabular} \\ \hline
Single-head                                                           & 1 M           & \begin{tabular}[c]{@{}c@{}}0.56\\ (24.89 \%)\end{tabular}  & \begin{tabular}[c]{@{}c@{}}\textbf{0.435}\\ ($\pm$ 0.139)\end{tabular} \\ \hline
\begin{tabular}[c]{@{}l@{}}Single-head \\ (Reduced)\end{tabular}      & \textbf{0.1 M}         & \begin{tabular}[c]{@{}c@{}}0.33\\ (14.67 \%)\end{tabular}  & \begin{tabular}[c]{@{}c@{}}0.534\\ ($\pm$ 0.205)\end{tabular} \\ \hline
\end{tabular}
\begin{tablenotes}[flushleft] \footnotesize
     \item $^*$ Percentage in the "Computation time" indicates the ratio of time spent on cost-critic versus others (e.g., simulation), where others are identical for all three methods.
    \end{tablenotes}
\label{table:multi_head_eval}
\end{table}

The results are shown in Figure \ref{fig:ablation}-B and Table \ref{table:multi_head_eval}. The multi-head method requires fewer trainable parameters than the single-head method. The multi-head cost value function requires 0.1 million trainable parameters, whereas single-head cost value functions require 1 million. Despite having considerably fewer neural network parameters to optimize, the policy optimized with the multi-head version yielded a similar reward as the single-head version (Figure \ref{fig:ablation}-B). This indicates that sharing the neural network backbone is permissible for each cost-value function, and the network capacity is sufficient to model them. Because only the output layer size increases in proportion to the number of constraints, the suggested technique is highly scalable and effective for solving problems with numerous constraints.

Additionally, the multi-head version showed the least computation time among the three methods due to its parallelized cost value function training and cost advantage computation (Table \ref{table:multi_head_eval}). The method's strength in terms of computation time grows as the number of constraints increases because, unlike other approaches, multi-head computation time does not increase linearly. The computation time for the cost value function training and cost advantage computation is not ignorable for all three methods, considering the relative ratio of time spent on other processes (e.g., simulation), and can be a bottleneck, especially for the single-head method, when the number of constraints increases. If we reduce the size of the cost value network, as done in the \textit{single-head (reduced)} method, the computation time problem can be alleviated. However, insufficient network capacity deteriorates the cost value prediction accuracy, exhibiting high value prediction loss with high variance. As the number of constraints increases, this becomes greater, affecting the derivation of an appropriate policy gradient from the computed cost advantage value.

\subsubsection{Algorithm hyperparameters}
We analyzed the sensitivity for two hyperparameter types: $t$ which determines the gradient profile of the logarithmic barrier functions, and $\alpha$ which determines how much the feasible region will be enlarged for the adaptive constraint thresholding. The results are illustrated in Figure \ref{fig:ablation}-C. If the barrier function was set to have gradient penalties concentrated near the threshold (i.e., large $t$) or the feasible region was enlarged too much (i.e., large $\alpha$), the barrier function applied small policy gradients to guide the policy toward the intended region, resulting in some constraints to be violated after the policy convergence (Table \ref{tab:contraint_satisfaction}). If the barrier function was modified to be far from an indicator function and overall have a smooth gradient profile (i.e., small $t$), the policy was trained to satisfy constraints with large margins, resulting in a low task reward acquisition (in our case, command tracking reward). If the feasible region was enlarged very slightly (i.e., small $\alpha$), it showed a small policy update step during the feasibility checking by the line search of the TRPO optimizer, especially at the initial training stage when the policy's exploration ratio is high. This can make the training slow depending on the policy exploration method and the training environment. 

Although the training characteristics change depending on the hyperparameters $t$ and $\alpha$, the trained results' variances were not very large in terms of constraint satisfaction, even though we modified hyperparameters on a large scale of ten. Furthermore, for all robots and various constraint designs used in our experiment, only a single set of hyperparameters, $t=100$ and $alpha=0.02$, was used and resulted in reliable constraint satisfaction and superior reward acquisition. This shows that the hyperparameters are robust across different RL environments once the suitable value is selected.

\section{Conclusion and Future Work}\label{sect:conclusion_future_work}
We proposed a learning framework for training neural network controllers for complex robotic systems consisting of both rewards and constraints. Suitable constraint types were suggested, where each can be configured in a way that best reflects the engineer's intent. An efficient policy optimization algorithm was then proposed, based on the previous works on constrained reinforcement learning literature, to search for a policy that maximizes the reward while satisfying multiple constraints. Our learning framework was applied for training controllers for legged robot locomotion in challenging terrains, which has previously been done with considerable effort in reward engineering. Extensive simulation and real-world experiments with diverse robots, possessing different morphologies and physical properties, showed the generalizability and capability of using constraints for training performant controllers with significantly less reward engineering. Further analysis of \textit{objective sensitivity} and \textit{constraint satisfaction} confirms that, in comparison to employing only rewards, the proposed framework can make the engineering process of generating desired motions more straightforward and efficient. To the best of our knowledge, our work is the first to show highly robust real-world robot performance for multiple legged robots using constrained reinforcement learning and demonstrate the benefits of constraints from an engineering standpoint.

We strongly believe that our work suggests a new direction for training neural network controllers for robotic systems. We provide a new perspective and capability of leveraging constraints in the learning pipeline, which replaces the substantial and laborious reward engineering that was unavoidable in the reward-only approach. Promising directions for future works include developing novel constraint formulations with different kernel functions based on specific use cases. It will also be fascinating to see how the learning framework is applied to diverse tasks and robot platforms. Lastly, leveraging several advanced exploration methods for training complex motions, rather than the basic random sampling from the Gaussian distribution, is promising in terms of making the training more data-efficient.

\section*{Appendix} \label{sec:appendix}
\subsection{Comparison of velocity command tracking performance with the reward-only framework}
As shown in Figure \ref{fig:reward_only_comparison}-A, our approach performed worse at velocity command tracking than the policy trained using the reward-only framework for the Raibo robot. This is due to the different kernel functions used for the command tracking reward. In our method, a quadratic function was used for simplicity, where the command tracking reward is defined as $r_c=-k_c (\vert\vert cmd_{v_{xy}}-V_{xy}\vert\vert^2 + (cmd_{w_{z}}-w_{z})^2) $. In the reward-only framework, an exponential function was used following the previous work \cite{ji2022concurrent, choi2023deformable}, where the command tracking reward is defined as $r_c=k_c (exp(-\vert\vert cmd_{v_{xy}}-V_{xy}\vert\vert^2) + exp(-1.5(cmd_{w_{z}}-w_{z})^2)) $.

\begin{table}[H]
\vspace*{-0.25cm}
\caption{Velocity command tracking performance comparison}
\centering
\begin{tabular}{|l|c|c|}
\hline
                    & Velocity command error & Torque \\ \hline
Ours + L2 (Raibo)  & 0.498 ($\pm$ 0.308)    & \textbf{24.769} ($\pm$ 5.811)                 \\ \hline
Ours + Exp (Raibo) & 0.367 ($\pm$ 0.290)    & 29.489 ($\pm$ 8.311)                 \\ \hline
Ours + Exp (Anymal C) & 0.547 ($\pm$ 0.380)    & 60.310 ($\pm$ 14.978)                 \\ \hline
Reward-only (Raibo)         & \textbf{0.359} ($\pm$ 0.354)    & 30.773 ($\pm$ 10.181)                \\ \hline
\end{tabular}
\label{table:command_tracking_eval}
\vspace*{-0.25cm}
\end{table}

\begin{figure}[h!]
\vspace*{-0.25cm}
\centering
\includegraphics[width=\columnwidth]{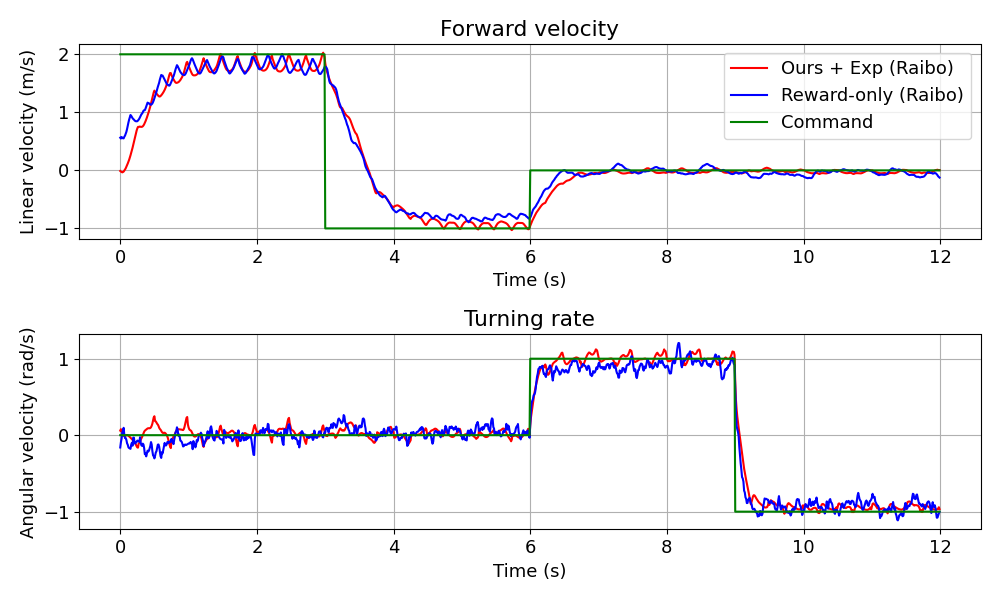}
\vspace*{-0.75cm}
\caption{Maximum velocity command tracking results}
\label{fig:velocity_plot}
\vspace*{-0.25cm}
\end{figure}

As shown in Table \ref{table:command_tracking_eval}, we can achieve the same command tracking performance with similar torque usage within our framework by applying the same exponential function for the command tracking reward and adjusting the reward coefficients accordingly ($k_c=200$, $\hat{k}_{\tau}=0.03$). When compared to a quadratic function, an exponential function has a steeper gradient near zero, resulting in more sensitive reward feedback for different command tracking errors. As a result, the robot can be trained to exhibit smaller command tracking errors. Additionally, similar maximum speeds can be achieved when given the maximum forward velocity (-1 \si{m/s} $\sim$ 2 \si{m/s}) and turning rate (-1 \si{rad/s} $\sim$ 1 \si{rad/s}) commands (Figure \ref{fig:velocity_plot}). 

To verify the robot generalization ability with different reward designs, we re-trained the controller for the Anymal C robot with the same reward coefficients used for the Raibo robot but with an exponential reward kernel (Table \ref{table:command_tracking_eval}). Similar results with Figure \ref{fig:reward_only_comparison}-A were obtained, indicating the maintenance of the robot generalization capability. This is because the critical component for achieving robot generalizability is the incorporation of constraints rather than specific reward designs. The proposed method offers strong robot generalizability as long as constraints are introduced and the effect is regardless of reward designs.

\subsection{Extra demonstration of constraint’s engineering advantages}
\begin{table*}[ht!]
\caption{Engineering scenario comparison}
\centering
\begin{tabular}{|l|c|c|cccc|l|}
\hline
\multirow{3}{*}{} & \multirow{3}{*}{Iteration} & \multirow{3}{*}{Parameters} & \multicolumn{4}{c|}{Evaluation} & \multicolumn{1}{c|}{\multirow{3}{*}{Comment}} \\ \cline{4-7}
 &  &  & \multicolumn{1}{c|}{Task} & \multicolumn{3}{c|}{Style / Safety} & \multicolumn{1}{c|}{} \\ \cline{4-7}
 &  &  & \multicolumn{1}{c|}{\begin{tabular}[c]{@{}c@{}}Velocity\\ reward\end{tabular}} & \multicolumn{1}{c|}{\begin{tabular}[c]{@{}c@{}}Body \\ contact\end{tabular}} & \multicolumn{1}{c|}{\begin{tabular}[c]{@{}c@{}}Contact \\ velocity\end{tabular}} & \begin{tabular}[c]{@{}c@{}}Gait\\ difference\end{tabular} & \multicolumn{1}{c|}{} \\ \hline
\begin{tabular}[c]{@{}l@{}}Initial \\ configuration\end{tabular} &  & \begin{tabular}[c]{@{}c@{}}$k_v=0.3$, \\ $k_{\tau}=4e-5$\end{tabular} & \multicolumn{1}{c|}{0.94} & \multicolumn{1}{c|}{0.32} & \multicolumn{1}{c|}{1.10} & 0.50 & \begin{tabular}[c]{@{}l@{}}- Too much \textcolor{ForestGreen}{body contact}\\ - Too much \textcolor{ForestGreen}{foot slippage}\\ - Not following the desired \textcolor{ForestGreen}{gait}\end{tabular} \\ \hline
\multirow{9}{*}{Reward-only} & 1 & $k_{bc}=1$ & \multicolumn{1}{c|}{0.95} & \multicolumn{1}{c|}{\textcolor{red}{0.21}} & \multicolumn{1}{c|}{0.96} & 0.52 & \begin{tabular}[c]{@{}l@{}}(A) Set \textcolor{blue}{body contact reward}\\ (R) Still \textcolor{red}{too much} body contact\end{tabular} \\ \cline{2-8} 
 & 2 & $k_{bc}=1e+2$ & \multicolumn{1}{c|}{0.94} & \multicolumn{1}{c|}{\textcolor{red}{0.02}} & \multicolumn{1}{c|}{2.10} & 0.50 & \begin{tabular}[c]{@{}l@{}}(A) Set a \textcolor{blue}{higher} body contact reward coefficient\\ (R) Still \textcolor{red}{frequent} body contact\end{tabular} \\ \cline{2-8} 
 & 3 & $k_{bc}=1e+4$ & \multicolumn{1}{c|}{0.96} & \multicolumn{1}{c|}{\textcolor{blue}{0.00}} & \multicolumn{1}{c|}{1.23} & 0.49 & \begin{tabular}[c]{@{}l@{}}(A) Set a \textcolor{blue}{higher} body contact reward coefficient\\ (R) Nearly zero body contact\\ (R) Move on to reducing foot slippage\end{tabular} \\ \cline{2-8} 
 & 4 & \begin{tabular}[c]{@{}c@{}}$k_{bc}=1e+4$, \\ $k_{cv}=1$\end{tabular} & \multicolumn{1}{c|}{0.95} & \multicolumn{1}{c|}{0.00} & \multicolumn{1}{c|}{\textcolor{red}{1.28}} & 0.50 & \begin{tabular}[c]{@{}l@{}}(A) Set \textcolor{blue}{slippage reward}\\ (R) Still \textcolor{red}{too much} foot slippage\end{tabular} \\ \cline{2-8} 
 & 5 & \begin{tabular}[c]{@{}c@{}}$k_{bc}=1e+4$, \\ $k_{cv}=1e+2$\end{tabular} & \multicolumn{1}{c|}{0.96} & \multicolumn{1}{c|}{0.00} & \multicolumn{1}{c|}{\textcolor{red}{0.17}} & 0.50 & \begin{tabular}[c]{@{}l@{}}(A) Set a \textcolor{blue}{higher} slippage reward coefficient\\ (R) Still \textcolor{red}{frequent} foot slippage\end{tabular} \\ \cline{2-8} 
 & 6 & \begin{tabular}[c]{@{}c@{}}$k_{bc}=1e+4$, \\ $k_{cv}=1e+4$\end{tabular} & \multicolumn{1}{c|}{\textbf{\textcolor{red}{0.78}}} & \multicolumn{1}{c|}{0.00} & \multicolumn{1}{c|}{0.05} & 0.50 & \begin{tabular}[c]{@{}l@{}}(A) Set a \textcolor{blue}{higher} slippage reward coefficient\\ (R) Acceptable foot slippage. \\ (R) However, \textcolor{red}{\textbf{degradation} in task performance}.\end{tabular} \\ \cline{2-8} 
 & 7 & \begin{tabular}[c]{@{}c@{}}$k_{bc}=1e+4$, \\ $k_{cv}=1e+3$\end{tabular} & \multicolumn{1}{c|}{0.92} & \multicolumn{1}{c|}{0.00} & \multicolumn{1}{c|}{\textcolor{blue}{0.09}} & 0.50 & \begin{tabular}[c]{@{}l@{}}(A) Set a \textcolor{blue}{smaller} slippage reward coefficient\\ (R) Acceptable slippage\\ (R) Move on to tracking the desired gait\end{tabular} \\ \cline{2-8} 
 & 8 & \begin{tabular}[c]{@{}c@{}}$k_{bc}=1e+4$, \\ $k_{cv}=1e+3$, \\ $k_{gp}=1e+2$\end{tabular} & \multicolumn{1}{c|}{0.91} & \multicolumn{1}{c|}{0.00} & \multicolumn{1}{c|}{0.12} & \textcolor{red}{0.49} & \begin{tabular}[c]{@{}l@{}}(A) Set \textcolor{blue}{gait reward}\\ (R) \textcolor{red}{Not tracking} the desired gait\end{tabular} \\ \cline{2-8} 
 & 9 & \begin{tabular}[c]{@{}c@{}}$k_{bc}=1e+4$, \\ $k_{cv}=1e+3$, \\ $k_{gp}=1e+4$\end{tabular} & \multicolumn{1}{c|}{0.87} & \multicolumn{1}{c|}{0.00} & \multicolumn{1}{c|}{\textbf{\textcolor{red}{0.12}}} & \textcolor{blue}{0.03} & \begin{tabular}[c]{@{}l@{}}(A) Set a \textcolor{blue}{higher} gait reward coefficient\\ (R) Acceptable gait tracking performance \\ (R) However, \textcolor{red}{slight \textbf{increase} in foot slippage}.\end{tabular} \\ \hline
\multirow{3}{*}{\textbf{\begin{tabular}[c]{@{}l@{}}Reward + \\ Constraints \\ (Ours)\end{tabular}}} & 1 & $D_{bc}=0.025$ & \multicolumn{1}{c|}{0.96} & \multicolumn{1}{c|}{\textcolor{blue}{0.00}} & \multicolumn{1}{c|}{1.44} & 0.51 & \begin{tabular}[c]{@{}l@{}}(A) Set \textcolor{blue}{body contact constraint} with \textcolor{blue}{near-zero} threshold\\ (R) Nearly zero body contact\\ (R) Move on to reducing foot slippage\end{tabular} \\ \cline{2-8} 
 & 2 & \begin{tabular}[c]{@{}c@{}}$D_{bc}=0.025$, \\ $D_{cv}=0.1$\end{tabular} & \multicolumn{1}{c|}{0.95} & \multicolumn{1}{c|}{0.00} & \multicolumn{1}{c|}{\textcolor{blue}{0.05}} & 0.50 & \begin{tabular}[c]{@{}l@{}}(A) Set \textcolor{blue}{slippage constraint} with \textcolor{blue}{0.1 \si{m/s}} threshold\\ (R) Acceptable slippage\\ (R) Move on to tracking the desired gait\end{tabular} \\ \cline{2-8} 
 & 3 & \begin{tabular}[c]{@{}c@{}}$D_{bc}=0.025$, \\ $D_{cv}=0.1$, \\ $D_{gp}=0.025$\end{tabular} & \multicolumn{1}{c|}{0.92} & \multicolumn{1}{c|}{0.00} & \multicolumn{1}{c|}{0.09} & \textcolor{blue}{0.024} & \begin{tabular}[c]{@{}l@{}}(A) Set \textcolor{blue}{gait pattern constraint} with \textcolor{blue}{near-zero} threshold\\ (R) Acceptable gait tracking performance\end{tabular} \\ \hline
\end{tabular}
\label{table:engineering_scenario_comparison}
\begin{tablenotes}[flushleft] \footnotesize
 \item $^*$ "(A)" indicates the engineer's action. "(R)" indicates the engineer's response after reviewing the trained results for the action taken in "(A)".
\end{tablenotes}
\vspace*{-0.25cm}
\end{table*}

The engineering benefits of constraints over rewards are demonstrated further with a simple toy example and a thorough comparison of the entire engineering scenario. Starting from a simple reinforcement learning setup with just two rewards, the goal is to train a neural network controller that exhibits a trotting gait pattern. The defined two rewards are designed for the robot to move forward with a velocity higher than 1m/s with minimal torque usage. The toy example setting is taken from the RaiSim \cite{hwangbo2018raisim} tutorial and the reward coefficients are also set the same. The complete engineering scenario is provided in Table \ref{table:engineering_scenario_comparison}.

When the robot is trained with just these two rewards, it exhibits three problems below.
\begin{itemize}
    \item Too much undesirable contact (e.g., self-collision, shank contact).
    \item Too much foot slippage.
    \item Not exhibiting the desired gait pattern (i.e., trot).
\end{itemize}

When constraints can be used during the training process, the above three problems can be solved easily by defining constraints appropriately. Based on the characteristics of each constraint type that we explained above, we defined three constraints: body contact constraint (probabilistic constraint, threshold=0.025), contact velocity constraint (average constraint, threshold=0.1 m/s), and gait pattern constraint (probabilistic constraint, threshold=0.025). By adding just these three constraints, we successfully trained a trotting controller with the above three problems solved. The video at \href{https://youtu.be/M-H1z2QRLeQ}{https://youtu.be/M-H1z2QRLeQ} shows how the policy evolves by introducing constraints one at a time. Similar controllers can be obtained by defining the cost functions as rewards and assigning appropriate reward coefficients to each of them. However, because reward coefficients can only model the relative relationships between each term and the trained result for a specific reward coefficient set is unpredictable, it takes several trials and errors to find the best coefficient set that can lead to the desired result. Furthermore, these coefficients must be carefully tuned so that they do not interfere with the main task performance (i.e., velocity reward). We believe that the corresponding experiment showcases how using constraints with rewards can make the entire engineering process of representing the engineer's intent more straightforward and effective. 

\ifCLASSOPTIONcaptionsoff
  \newpage
\fi


\typeout{}
\bibliographystyle{bibtex/IEEEtran}
\bibliography{bibtex/IEEEabrv,bibtex/library}


\begin{IEEEbiography}[{\includegraphics[width=1in,height=1.25in,clip,keepaspectratio]{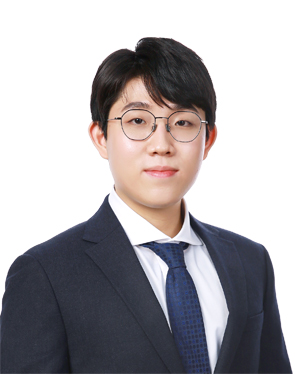}}]{Yunho Kim}
is a Master student at the Robotics and Artificial Intelligence Lab, Korea Advanced Institute of Science and Technology~(KAIST), under the supervision of Prof. J. Hwangbo. His research interests are legged robot locomotion and navigation. He received his B.Sc. in Mechanical Engineering from Seoul National University, Republic of Korea, in 2022.
\end{IEEEbiography}

\begin{IEEEbiography}[{\includegraphics[width=1in,height=1.25in,clip,keepaspectratio]{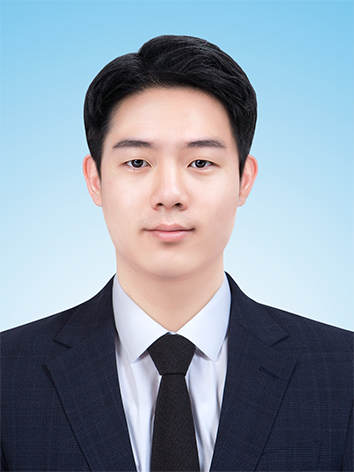}}]{Hyunsik Oh}
is a Master student at the Robotics and Artificial Intelligence Lab, Korea Advanced Institute of Science and Technology~(KAIST), under the supervision of Prof. J. Hwangbo. His research interests include robot design and learning-based control. He received his B.Sc. in Mechanical Engineering from Hanyang University, Republic of Korea, in 2023
\end{IEEEbiography}

\begin{IEEEbiography}[{\includegraphics[width=1in,height=1.25in,clip,keepaspectratio]{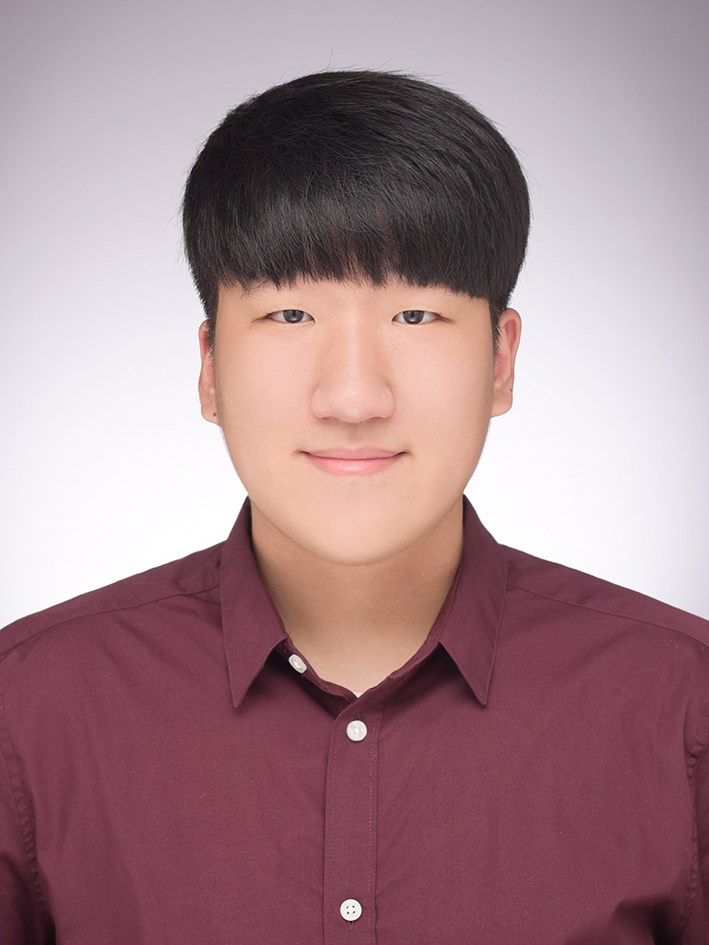}}]{Jeonghyun Lee}
is a Master student at the Robotics and Artificial Intelligence Lab, Korea Advanced Institute of Science and Technology~(KAIST), under the supervision of Prof. J. Hwangbo. His research interests are autonomous vehicle navigation and deep learning methods. He received his B.Sc. in Mechanical Engineering from KAIST, Republic of Korea, in 2022.
\end{IEEEbiography}

\begin{IEEEbiography}[{\includegraphics[width=1in,height=1.25in,clip,keepaspectratio]{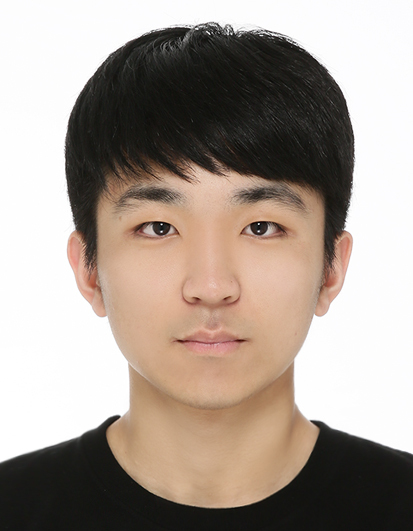}}]{Jinhyeok Choi}
is an Undergraduate research student
at the Robotics and Artificial Intelligence Lab,
Korea Advanced Institute of Science and Technology~(KAIST), under the supervision of Prof. J. Hwangbo.
His research interests are legged robot actuators
and locomotion.
\end{IEEEbiography}

\begin{IEEEbiography}[{\includegraphics[width=1in,height=1.25in,clip,keepaspectratio]{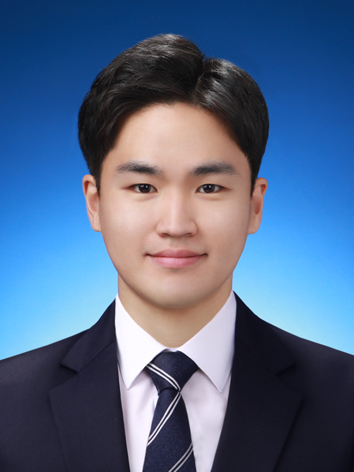}}]{Gwanghyeon Ji}
is a PhD student at the Robotics and Artificial Intelligence Lab, Korea Advanced Institute of Science and Technology~(KAIST), under the supervision of Prof. J. Hwangbo. His main research interest is legged robot locomotion. He received his B.Sc. in Mechanical Engineering from Pohang University of Science and Technology, Republic of Korea, in 2017 and M.Sc. in Mechanical Engineering from KAIST, in 2023.
\end{IEEEbiography}

\begin{IEEEbiography}[{\includegraphics[width=1in,height=1.25in,clip,keepaspectratio]{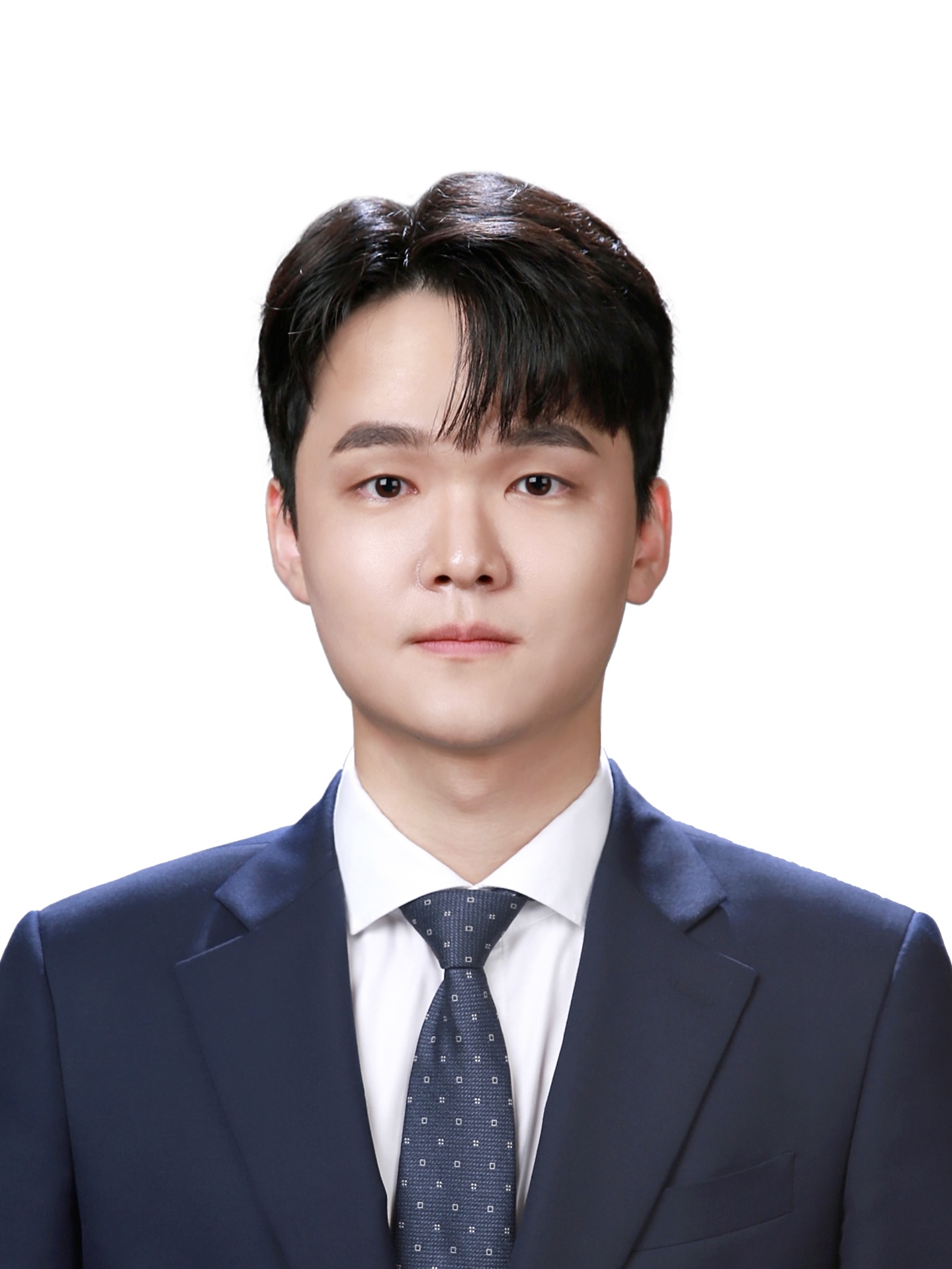}}]{Moonkyu Jung}
is a Master student at the Robotics and Artifical Intelligence Lab, Korea Advanced Institute of Science and Technology~(KAIST), under the supervision of Prof. J. Hwangbo. His research interests are legged locomotion and loco-manipulation. He received his B.Sc in Mechanical Engineering from KAIST, Republic of Korea, in 2023.
\end{IEEEbiography}

\begin{IEEEbiography}[{\includegraphics[width=1in,height=1.25in,clip,keepaspectratio]{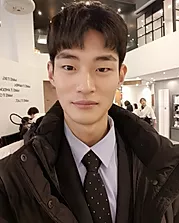}}]{Donghoon Youm}
received the B.Eng. and M.Eng. degrees in Mechanical Engineering from Korea Advanced Institute of Science and Technology~(KAIST) in 2020 and 2022, respectively. He is currently pursuing a PhD degree at Robotics and Artificial Intelligence Lab in the Department of Mechanical Engineering, KAIST. His research interests include imitation learning, reinforcement learning, state estimation, and system identification for legged robots.
\end{IEEEbiography}

\begin{IEEEbiography}[{\includegraphics[width=1in,height=1.25in,clip,keepaspectratio]{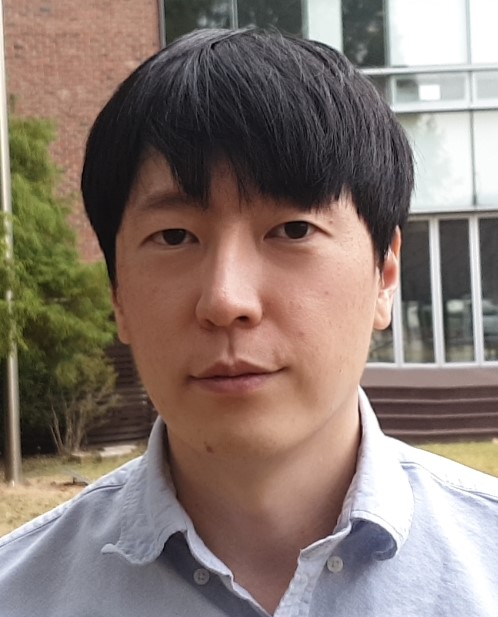}}]{Jemin Hwangbo}
 serves as an assistant professor in the Department of Mechanical Engineering at KAIST and concurrently holds the position of Director at the Robotics and Artificial Intelligence Lab (RAI Lab). His group's research is primarily centered around legged robotics, encompassing areas such as design, vision, control, and navigation. He obtained his B.S. degree from the University of Toronto in 2006. His M.S. and Ph.D. were conferred by ETH Zurich in 2013 and 2019, respectively.

\end{IEEEbiography}






\end{document}